\pdfoutput=1

\documentclass[11pt]{article}

\usepackage[]{ACL2023}

\usepackage{times}
\usepackage{latexsym}

\usepackage[T1]{fontenc}

\usepackage[utf8]{inputenc}

\usepackage{microtype}

\usepackage{inconsolata}

\usepackage{microtype}
\usepackage{times}
\usepackage{latexsym}
\usepackage{booktabs} 
\usepackage{multirow}
\usepackage{url}
\usepackage{amsmath}
\usepackage{bm}
\usepackage{amsfonts}
\usepackage{graphicx}
\usepackage{diagbox}
\usepackage{xspace}
\usepackage{courier}
\usepackage{mathrsfs}
\usepackage{appendix}
\usepackage{float}
\usepackage{framed} 
\usepackage{color}
\usepackage{xspace}
\usepackage{amsfonts}
\usepackage{amssymb}
\usepackage{diagbox}

\usepackage{amsmath}

\usepackage{microtype}
\usepackage{textcomp}
\usepackage{amsfonts}
\usepackage{amssymb}
\usepackage{diagbox}
\usepackage{algorithm}
\usepackage{algorithmic}

\usepackage{makecell}
\usepackage{xcolor}
\usepackage{pgfplots}
\definecolor{tiffanyblue}{RGB}{129,216,208}
\definecolor{bangdiblue}{RGB}{0,149,182}
\definecolor{kleinblue}{RGB}{0,47,167}
\definecolor{purple}{RGB}{138,43,226}
\usepackage{amssymb}
\usetikzlibrary{shapes}
\usepackage{circledtext}
\usetikzlibrary{arrows,decorations.pathmorphing,backgrounds,positioning,fit,petri}
\usetikzlibrary{arrows.meta,fit,shapes.arrows}
\usepackage{caption} 
\usepackage{xcolor}
\usepackage{pgfplots}
\usepackage{tikz}
\usepackage{subfig}
\usetikzlibrary{positioning,shapes,arrows,shadows,patterns}
\usepackage{tkz-kiviat,pgfplots}
\pgfplotsset{compat=newest}

\title{Augmenting Large Language Model Translators via Translation Memories}

\author{Yongyu Mu\textsuperscript{1}\thanks{\xspace\xspace Equal contribution.}, Abudurexiti Reheman\textsuperscript{1}\footnotemark[1], Zhiquan Cao\textsuperscript{1}, Yuchun Fan\textsuperscript{1}, \\
{\bf Bei Li\textsuperscript{1}, Yinqiao Li\textsuperscript{1}, Tong Xiao\textsuperscript{1,2}\thanks{\xspace\xspace Corresponding author.}, Chunliang Zhang\textsuperscript{1,2} \and Jingbo Zhu\textsuperscript{1,2}} \\
	\textsuperscript{1}NLP Lab, School of Computer Science and Engineering, \\ Northeastern University, Shenyang, China\\
	\textsuperscript{2}NiuTrans Research, Shenyang, China\\
	\ttfamily{lixiaoyumu9@gmail.com rexiti\_neu@outlook.com}\\
	\ttfamily{\{xiaotong,zhujingbo\}@mail.neu.edu.cn}
}

\begin{document}
\maketitle

\begin{abstract}

Using translation memories (TMs) as prompts is a promising approach to in-context learning of machine translation models. In this work, we take a step towards prompting large language models (LLMs) with TMs and making them better translators. We find that the ability of LLMs to ``understand'' prompts is indeed helpful for making better use of TMs. Experiments show that the results of a pre-trained LLM translator can be greatly improved by using high-quality TM-based prompts. These results are even comparable to those of the state-of-the-art NMT systems which have access to large-scale in-domain bilingual data and are well tuned on the downstream tasks.

\end{abstract}

\section{Introduction}

Marrying the world of translation memory (TM) and the world of neural machine translation (NMT) is a challenging but interesting problem in natural language processing (NLP). Previous work along this line of research either requires architecture changes of NMT models and/or additional training \cite{gu2018search, bulte2019neural, xu2020boosting, hossain2020simple, he2021fast} or constructing translation knowledge base from TM \cite{zhang2018guiding, khandelwal2020nearest,meng2021fast}.

More recently, researchers have been aware of the strength of prompting techniques for one-shot/few-shot machine translation \cite{vilar2022prompting,DBLP:journals/corr/abs-2212-02437,zhang2023prompting}. In particular, \citet{reheman2023prompting} investigated one-shot learning methods for NMT by simply viewing TMs as prompts. The result of their work is a stronger NMT system that works in the same way as usual but can be prompted when TMs are available. Interestingly, they found that the ability of NMT models to ``understand'' prompts plays an important role in this type of system. Prompts are still difficult to use if NMT systems are weak.

\begin{table}[t!]
\small
\resizebox{0.49\textwidth}{!}{
\centering
\begin{tabular}{l|ccc}
\toprule
\textbf{Method} & \textbf{w/o-arch-change} & {\textbf{w/o-base}}  & \textbf{few-shot} \\
\midrule
\citet{zhang2018guiding} & yes &  &  \\
\citet{he2021fast} &  & yes &  \\
\citet{khandelwal2020nearest} & yes & &  \\
\citet{reheman2023prompting} & yes & yes & one-shot \\
\textsc{TMPlm} (our) & yes & yes & yes \\
\bottomrule
\end{tabular}
}
 \caption{Methods of using TM for better MT. w/o-arch-change = without architecture changes or training, w/o-base = without constructing translation knowledge base from TM, and few-shot = few-shot learning.}
\vspace{-1.1em}
 \label{tab:t1}
\end{table} 

In this work, we take a step forward. We treat large language models (LLMs) as machine translation systems and prompt them with TMs (see Table \ref{tab:t1} for a comparison of different methods). This is in part motivated by recent developments of LLMs: one of the most powerful properties of LLMs is their ability to understand and respond to complex instructions and questions \cite{ouyang2022training,thoppilan2022lamda}. We show that this ability is crucial for in-context learning of TM-based prompts, and LLM-based translation systems can be greatly improved by using simple instruction-like prompts. To this end, we propose \textbf{T}ranslation \textbf{M}emory \textbf{P}rompting for large \textbf{L}anguage \textbf{M}odels, namely \textsc{TMPlm} - a simple but effective approach to injecting TMs into LLM translators.

We experiment with our method on a GPT-based LLM (\texttt{text-davinci-003}\footnote{We will refer to it as \texttt{davinci-003} later in the paper.}). On translation tasks ranging over multiple languages and domains, TM-based prompting improves the LLM-based translation system by 20 to 30 BLEU points, showing better performance than a well-tuned, large-scale, in-domain NMT system on most of the tasks. We also compare different kinds of prompt templates and discuss some interesting issues, such as the role of prompting in treating LLMs as translators.

\begin{figure*}
\centering
\begin{tikzpicture}
\useasboundingbox (-10em,-10em) rectangle (10em,3em);
\tikzstyle{modelnode} = [rectangle,draw=none,rounded corners=2pt,inner sep=2pt,minimum height=4.5em,minimum width=2em,font=\small,anchor=north,rotate=90]
\node [draw=black,inner sep=0.4em,rounded corners=4pt,thick,fill=gray!10,minimum width=41.5em,minimum height=7.6em] (it) at (0,0){};
\node [draw=black,inner sep=0.4em,rounded corners=4pt,thick,fill=gray!10,minimum width=41.5em,minimum height=5.5em] (it_1) at ([xshift=0em,yshift=-3em]it.south){};
\node (b)[anchor=north,align=center] at ([xshift=-0.9em,yshift=-0.5em]it.north){$f(\cdot)$\hspace{0.87em}: What is the translation of "\hspace{1em}" from \textit{src-lang} to \textit{tgt-lang}? Only translation results};
\node (b_1) [anchor=south,align=center] at ([xshift=-12.1em,yshift=-1.2em]b.south){are required.};
\node (b_2) [anchor=south,align=center,fill=red!20,rounded corners=2pt,inner sep=2.5pt] at ([xshift=-3.41em,yshift=-1.33em]b.north){$\mathbf{x}$};
\node [modelnode,align=center] (a) at ([xshift=-22em,yshift=-4em]it.north){\ttfamily{\textbf{INSTRUCTION}}};
\node (c) [anchor=south,align=center] at([xshift=-0.62em,yshift=-4.8em]it.north){$f_{\mathrm{ref}}(\cdot)$: If the translation of "\hspace{1.7em}" from \textit{src-lang} to \textit{tgt-lang} is "\hspace{1.7em}" and the translation of};
\node(c1) [anchor=south,align=center,fill=blue!20,,rounded corners=2pt,inner sep=1pt] at ([xshift=-5.75em,yshift=-1.4em]c.north){$\mathbf{x}_{\mathrm{tm}}^{1}$};
\node(c1_1) [anchor=south,align=center,fill=green!20,,rounded corners=2pt,inner sep=1pt] at ([xshift=8.12em,yshift=-1.4em]c.north){$\mathbf{y}_{\mathrm{tm}}^{1}$};
\node (c1_1) [anchor=south,align=center] at ([xshift=1.49em,yshift=-1.25em]c.south){"\hspace{1.7em}" from \textit{src-lang} to \textit{tgt-lang} is "\hspace{1.7em}", then what is the translation of "\hspace{1em}" from};
\node (c1_2) [anchor=south,align=center] at ([xshift=-4.9em,yshift=-1.3em]c1_1.south){\textit{src-lang} to \textit{tgt-lang}? Only translation results are required.};
\node(c1_2) [anchor=south,align=center,fill=blue!20,,rounded corners=2pt,inner sep=1pt] at ([xshift=-13.73em,yshift=-2.7em]c.north){$\mathbf{x}_{\mathrm{tm}}^{2}$};
\node(c1_3) [anchor=south,align=center,fill=green!20,,rounded corners=2pt,inner sep=1pt] at ([xshift=0.14em,yshift=-2.7em]c.north){$\mathbf{y}_{\mathrm{tm}}^{2}$};
\node(c1_6) [anchor=south,align=center,fill=red!20,,rounded corners=2pt,inner sep=2.5pt] at ([xshift=14.9em,yshift=-2.65em]c.north){$\mathbf{x}$};
\node (d) [modelnode,align=center] at ([xshift=-4.5em,yshift=-6.7em]a.south){\ttfamily{\textbf{CODE}}};
\node (e) [anchor=south,align=center] at ([xshift=5.75em,yshift=0.6em]d.south){$f(\cdot)$\hspace{0.86em}: [\textit{src-lang}]=[\hspace{1.1em}] [\textit{tgt-lang}]=};
\node (e1) [anchor=south,align=center,fill=red!20,,rounded corners=2pt,inner sep=2.5pt] at ([xshift=1.46em,yshift=-1.3em]e.north) {$\mathbf{x}$};
\node (f) [anchor=south,align=center] at ([xshift=17.1em,yshift=-0.9em]d.south){$f_{\mathrm{ref}}(\cdot)$: [\textit{src-lang}]=[\hspace{1.9em}] [\textit{tgt-lang}]=[\hspace{1.9em}] [\textit{src-lang}]=[\hspace{1.9em}] [\textit{tgt-lang}]=[\hspace{1.9em}] [\textit{src-lang}]=};
\node (f) [anchor=south,align=center] at ([xshift=-12em,yshift=-1.5em]f.south){[\hspace{1.1em}] [\textit{tgt-lang}]=};
\node (f1) [anchor=south,align=center,fill=blue!20,rounded corners=2pt,inner sep=1pt] at ([xshift=2.51em,yshift=0.1em]f.north){$\mathbf{x}_{\mathrm{tm}}^{1}$};
\node (f2) [anchor=south,align=center,fill=green!20,rounded corners=2pt,inner sep=1pt] at ([xshift=9.76em,yshift=0.1em]f.north){$\mathbf{y}_{\mathrm{tm}}^{1}$};
\node (f3) [anchor=south,align=center,fill=blue!20,rounded corners=2pt,inner sep=1pt] at ([xshift=17.11em,yshift=0.1em]f.north){$\mathbf{x}_{\mathrm{tm}}^{2}$};
\node (f4) [anchor=south,align=center,fill=green!20,rounded corners=2pt,inner sep=1pt] at ([xshift=24.31em,yshift=0.1em]f.north){$\mathbf{y}_{\mathrm{tm}}^{2}$};
\node (f2) [anchor=south,align=center,fill=red!20,rounded corners=2pt,inner sep=2.5pt] at ([xshift=-2.35em,yshift=-1.26em]f.north){$\mathbf{x}$};
\end{tikzpicture} 
\caption{Two styles of template. $f(\cdot)$ denotes a template by which we represent the input sentence as the input of the translation model (such as LLM in this figure). $f_{\mathrm{ref}} (\cdot)$ is a new template involving outputs of a TM ($k=2$ in this example). $\mathbf{x}$ in red stands for the sentence that needs to be translated. $\mathbf{x}_{\mathrm{tm}}$ in blue and $\mathbf{y}_{\mathrm{tm}}$ in green stand for the source and target sentence found in the TM, respectively. Both \textit{src-lang} and \textit{tgt-lang} need to be replaced by the names of the source and target language.}
\label{fig:f1}
\vspace{-0.7em}
\end{figure*}
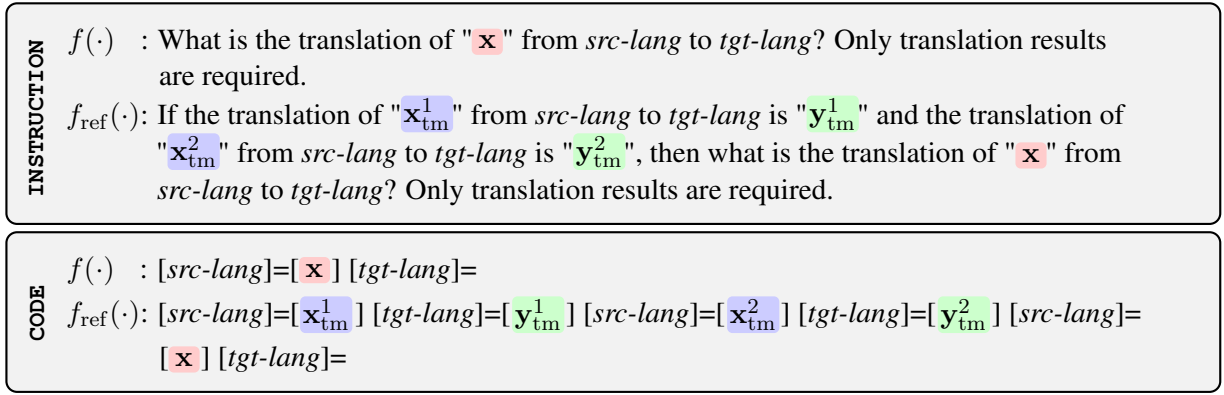

\section{Prompting Methods}
\label{sec:s1}
TM is a database that contains the bilingual translation history of professional translators. It is usually used to help the translation of the test sentence by providing similar sentence pairs, which may have translation hints, such as similar sentence patterns, phrases, lexicons, terminologies, or other translation knowledge. Either an NMT model or an LLM need to \textit{dig out} those hints and ignore the irrelevant content. This motivates us to investigate prompting LLMs with TMs benefiting from their dazzling ability of ``understand'' prompts.


Suppose we have a TM database that retains a collection of pairs of sentences. Given a source-language sentence $\mathbf{x}$, the database returns $k$ most similar sentences $\mathbf{X}_{\mathrm{tm}}=\{\mathbf{x}_{\mathrm{tm}}^{1},...,\mathbf{x}_{\mathrm{tm}}^{k}\}$ along with their corresponding translations $\mathbf{Y}_{\mathrm{tm}}=\{\mathbf{y}_{\mathrm{tm}}^{1},...,\mathbf{y}_{\mathrm{tm}}^{k}\}$. Now suppose we have a pre-trained translation model (either an NMT model or an LLM) that takes $\mathbf{x}$ in some way and outputs a translation $\mathbf{y}$, written as
\begin{eqnarray}
\mathbf{y} & = & \mathrm{Trans}(f(\mathbf{x}))
\end{eqnarray}

\noindent where $\mathrm{Trans}(\cdot)$ denotes the translation model, and $f(\cdot)$ denotes a template by which we represent $\mathbf{x}$ as the input of $\mathrm{Trans}(\cdot)$. For example, if $\mathrm{Trans}(\cdot)$ is an NMT model, $f(\mathbf{x}) = \mathbf{x}$; if $\mathrm{Trans}(\cdot)$ is a generative LLM, $f(\mathbf{x})$ could be an instruction involving $\mathbf{x}$.

We then wish to use this model to generate a new translation $\mathbf{y}'$ by considering $(\mathbf{X}_{\mathrm{tm}}, \mathbf{Y}_{\mathrm{tm}})$ as instances for reference. This can be written as
\begin{eqnarray}
\mathbf{y}' & = & \mathrm{Trans}(f_{\mathrm{ref}}(\mathbf{x},\mathbf{X}_{\mathrm{tm}},\mathbf{Y}_{\mathrm{tm}}))
\end{eqnarray}

\noindent Here $f_{\mathrm{ref}}(\mathbf{x},\mathbf{X}_{\mathrm{tm}},\mathbf{Y}_{\mathrm{tm}})$ is a new template involving $(\mathbf{X}_{\mathrm{tm}}, \mathbf{Y}_{\mathrm{tm}})$.

In this work, we focus on the case in which a powerful generative LLM (such as \texttt{ChatGPT}) is used to perform translation. The input of $\mathrm{Trans}(\cdot)$ could be an instruction or question-like text, and so we can design $f_{\mathrm{ref}}(\cdot)$ in many different ways. In Figure \ref{fig:f1},  we present two types of templates: the instruction-style template and the code-style template. These designs come from a consideration of the human instruction tuning and the code training used in developing \texttt{davinci-003}. For a more extensive discussion of template design, see Appendix \ref{app:dif_prompt_performance}.

It is worth emphasizing that, while we restrict ourselves to TM-based prompts in experiments, we can apply this general approach to deal with other knowledge about translation. As a simple example, we can extend $(\mathbf{X}_{\mathrm{tm}}, \mathbf{Y}_{\mathrm{tm}})$ to term or phrase translations. Also, when some MT systems are available, we can make use of automatic translations from other systems to define prompts.

\begin{table*}[t!]
\small
    \centering

 \resizebox{1.0\linewidth}{!}{
    \begin{tabular}{cc|cc|cc|ccc}
        \toprule
        \multicolumn{2}{c|}{\multirow{3}{*}[-0.1ex]{\normalsize{\textbf{Data}}}} & \multicolumn{2}{c|}{\normalsize{\textbf{WMT19 200M}}} & \multicolumn{2}{c|}{\normalsize{\textbf{WMT21 4B}}} & \multicolumn{3}{c}{\normalsize{\textbf{davinci-003 175B}}} \\
        & & \multicolumn{1}{c}{\multirow{2}{*}[-0.1ex]{\makecell{\textbf{NMT}}}} & \multicolumn{1}{c|}{\multirow{2}{*}[-0.1ex]{\makecell{\textbf{NMT+TM}}}} & \multicolumn{1}{c}{\multirow{2}{*}[-0.1ex]{\makecell{\textbf{NMT}}}} & \multicolumn{1}{c|}{\multirow{2}{*}[-0.1ex]{\makecell{\textbf{NMT+TM}}}} & \multicolumn{1}{c}{\multirow{2}{*}[0.4ex]{\textbf{LLM}}} &
        \multicolumn{1}{c}{\multirow{2}{*}[0.4ex]{\textbf{LLM+TM}}}  & \multicolumn{1}{c}{\multirow{2}{*}[0.4ex]{\textbf{LLM+TM}}}\\    
        & & & & & & \multicolumn{1}{c}{\multirow{1}{*}[0.1ex]{\footnotesize{(zero-shot)}}} &\multicolumn{1}{c}{\multirow{1}{*}[0.1ex]{\footnotesize{(one-shot)}}} & \multicolumn{1}{c}{\multirow{1}{*}[0.1ex]{\footnotesize{(few-shot)}}} \\
        \midrule
        \multirow{2}{*}{DGT-TM} & de $\rightarrow$ en & 45.40 & 54.03$_{(+8.63)}$ & 51.62 & 69.39$_{(+17.77)}$ & 38.89 & 66.90$_{(+28.01)}$ & \textbf{69.99}$_{(+31.10)}$ \\   
        & en $\rightarrow$ de & 39.03 & 44.77$_{(+5.74)}$ & 42.48 & 60.09$_{(+17.61)}$ & 29.00 & 57.39$_{(+28.39)}$ & \textbf{62.02}$_{(+33.02)}$ \\
        \midrule
        \multirow{2}{*}{JRC-A} & de $\rightarrow$ en & 45.90 & 50.95$_{(+5.05)}$ & 51.72 &  62.99$_{(+11.27)}$ & 40.75 & 62.23$_{(+21.48)}$ & \textbf{65.55}$_{(+24.80)}$ \\
        & en $\rightarrow$ de & 40.10 & 43.41$_{(+3.31)}$ & 41.71 & 56.21$_{(+14.50)}$ & 29.83 & 55.01$_{(+25.18)}$ & \textbf{57.30}$_{(+27.47)}$ \\
        \bottomrule
    \end{tabular} 
    }

    \caption{BLEU scores of NMT models and LLMs on the DGT-TM and JRC-A dataset. WMT19 200M indicates WMT19 champion models \cite{ng2019facebook}, containing 200 million parameters. WMT21 4B indicates WMT21 champion models \cite{tran2021facebook} trained by multi language-pairs data containing 4 billion parameters. One-shot and few-shot represent the results of \textsc{TMPlm} with $k=1$ and $k=5$, respectively. The BLEU improvements are reported in subscripts. See Table \ref{tab:app:main_experiments_by_COMET-22} for the COMET-22 version.
    }

    \label{tab:t2}
    
\vspace{-0.5em}

\end{table*}

\section{Experiments}

\subsection{Data and LLM Setup}

We tested our method (denoted by \textsc{TMPlm}) on three widely-used datasets of TM: DGT-TM \cite{Ralf2012dgt}, JRC-Acquis (JRC-A) \cite{steinberger2006jrc} and the multi-domain dataset described in \cite{aharoni2020unsupervised}. To ensure a fair comparison, we adopted the same preprocessing steps as outlined in \citet{reheman2023prompting} for data cleanup and training/testing data split.

For LLMs, we chose the \texttt{davinci-003} model developed by OpenAI because it is currently one of the state-of-the-art generative LLMs. The model was configured with default values of all parameters, except that the sampling temperature was set to 0. In the experiments, we used the code-style template and set $k$ to 5 by default. The quality of translations was mainly evaluated using \texttt{multi-bleu.perl} from Moses\footnote{\url{http://www.statmt.org/moses/}}. In addition, following the recommend of using neural network-based metrics in machine translation evaluation \cite{DBLP:conf/wmt/FreitagRMLSAKFLM22}, we also used COMET-22\footnote{\url{https://github.com/Unbabel/COMET}} (\textit{wmt22-COMET-da}) \cite{DBLP:conf/wmt/ReiSAZFGLCM22} to make a complementary evaluation. See more details about data processing in Appendixes \ref{app:data pre-process} and \ref{app:data post-process}.

\begin{figure}[t!]
\centering
\makeatletter
\def\tkz@KiviatGrad[#1](#2){%
\begingroup
\pgfkeys{/kiviatgrad/.cd,
graduation distance= 0 pt,
prefix ={},
suffix={},
unity=1,
label precision/.store in=\gradlabel@precision,
label precision=1,
zero point/.store in=\tkz@grad@zero,
zero point=0
}
\pgfqkeys{/kiviatgrad}{#1}%
\let\tikz@label@distance@tmp\tikz@label@distance
\global\let\tikz@label@distance\tkz@kiv@grad
 \foreach \nv in {0,...,\tkz@kiv@lattice}{
 \pgfmathsetmacro{\result}{\tkz@kiv@unity*\nv+20} 
 \protected@edef\tkz@kiv@gd{%
    \tkz@kiv@prefix%
    \pgfmathprintnumber[precision=\gradlabel@precision,fixed]{\result}
    \tkz@kiv@suffix} 
    \path[/kiviatgrad/.cd,#1] (0:0)--(360/\tkz@kiv@radial*#2:\nv*\tkz@kiv@gap)
       node[label=(360/\tkz@kiv@radial*#2):\tiny\tkz@kiv@gd] {}; 
      }
 \let\tikz@label@distance\tikz@label@distance@tmp  
\endgroup
}%

\def\tkz@KiviatLine[#1](#2,#3){%
\begingroup
\pgfkeys{/kiviatline/.cd,
fill= {},
opacity=.5,
zero point/.store in=\tkz@line@zero,
zero point=0
}
%
%
\pgfqkeys{/kiviatline}{#1}
\ifx\tkzutil@empty\tkz@kivl@fill \else 
\begin{scope}[on background layer]
 \path[fill=\tkz@kivl@fill,opacity=\tkz@kivl@opacity] (360/\tkz@kiv@radial*0:{(#2+\tkz@line@zero)*\tkz@kiv@gap*\tkz@kiv@step})   
\foreach \v [count=\rang from 1] in {#3}{%
 -- (360/\tkz@kiv@radial*\rang:{(\v+\tkz@line@zero)*\tkz@kiv@gap*\tkz@kiv@step}) } -- (360/\tkz@kiv@radial*0:{(#2+\tkz@line@zero)*\tkz@kiv@gap*\tkz@kiv@step}); 
 \end{scope}
 \fi       
\draw[#1,opacity=1,overlay] (0:{(#2+\tkz@line@zero)*\tkz@kiv@gap}) plot coordinates {(360/\tkz@kiv@radial*0:{(#2+\tkz@line@zero)*\tkz@kiv@gap*\tkz@kiv@step})}  
\foreach \v [count=\rang from 1] in {#3}{%
 -- (360/\tkz@kiv@radial*\rang:{(\v+\tkz@line@zero)*\tkz@kiv@gap*\tkz@kiv@step}) plot coordinates {(360/\tkz@kiv@radial*\rang:{(\v+\tkz@line@zero)*\tkz@kiv@gap*\tkz@kiv@step})}} -- (360/\tkz@kiv@radial*0:{(#2+\tkz@line@zero)*\tkz@kiv@gap*\tkz@kiv@step});   
\endgroup
}%

\makeatother
\definecolor{c1}{HTML}{81BECE}
\definecolor{c2}{HTML}{378BA4}
\tikzset{global scale/.style={
    scale=#1,
    every node/.append style={scale=#1}
  }
}
\begin{tikzpicture}[
  label distance=13em,
  global scale = 0.73,
  plot1/.style={
    thin,
    draw=c1,
    fill=c1,
    mark=square*,
    mark options={
     ball color=blue, 
     mark size=4pt
    },
    opacity=.5
  },
  plot2/.style={
    thin,
    draw=c2,
    fill=c2,
    mark=triangle*,
    mark options={
      mark size=4pt,
      ball color=blue
    },
    opacity=.5
  },
  plot3/.style={
    thin,
    draw=red!30,
    fill=red!30,
    mark=square*,
    mark options={
      mark size=4pt,
      ball color=yellow
    },
    opacity=.5
  },
  plot4/.style={
    thin,
    draw=red!50,
    fill=red!50,
    mark=triangle*,
    mark options={
      mark size=4pt,
      ball color=blue
    },
    opacity=.5
  }
]
\useasboundingbox (-25em,-5em) rectangle (10em,7em);
\begin{scope}[scale=0.36,local bounding box=a,shift={(-50em,0)}]

\newcommand\KivStep{0.1}
\pgfmathsetmacro\Unity{1/\KivStep}
\newcommand\zeroshift{20.00}

 \tkzKiviatDiagram[
   radial  style/.style ={-},
   rotate=90, 
   lattice style/.style ={black!30},
   step=\KivStep,
   gap=1,
   lattice=5,
]%
{cs$\rightarrow$en,cs$\rightarrow$it,,ro$\rightarrow$en,en$\rightarrow$cs,it$\rightarrow$cs,,en$\rightarrow$ro}
\node [align=center,rotate=90]at (0em,16em){de$\rightarrow$en};
\node [align=center,rotate=270]at (0em,-16em){en$\rightarrow$de};

\tkzKiviatLine[
  plot3
](38.58,27.93,39.74,39.66,32.58,27.22,33.56,30.18)
\tkzKiviatLine[
  plot4
](17.50,7.02,21.37,16.34,8.02,2.85,7.06,8.74)

\tkzKiviatGrad[unity=\Unity, label precision=2, zero point=\zeroshift](0) 
\end{scope}
\begin{scope}[scale=0.36,local bounding box=b,shift={(-15em,0)}]

\newcommand\KivStep{0.1}
\pgfmathsetmacro\Unity{1/\KivStep}
\newcommand\zeroshift{20.00}

 \tkzKiviatDiagram[
   radial  style/.style ={-},
   rotate=90,
   lattice style/.style ={black!30},
   step=\KivStep,
   gap=1,
   lattice=5
]%
{es$\rightarrow$en,fr$\rightarrow$en,,de$\rightarrow$fr,en$\rightarrow$es,en$\rightarrow$fr,,fr$\rightarrow$de}
\node [align=center,rotate=90]at (0em,16.5em){it$\rightarrow$en};
\node [align=center,rotate=270]at (0em,-16.5em){en$\rightarrow$it};
\tkzKiviatLine[
plot1
](41.25,44.45,41.80,33.52,39.18,44.60,37.71,30.08)
\tkzKiviatLine[
plot2
](21.89,24.06,21.13,14.78,17.10,23.78,13.53,8.93)
\tkzKiviatGrad[unity=\Unity, zero point=\zeroshift](0) 
\end{scope}

\end{tikzpicture}
\caption{Comparison of LLM w/o and w/ TMs (one-shot) on 8 language-pairs from JRC-A. Points in deep and light color stand for the BLEU scores of LLM w/o and w/ TM, respectively.}
\label{fig:f2}
\vspace{-1.1em}
\end{figure}
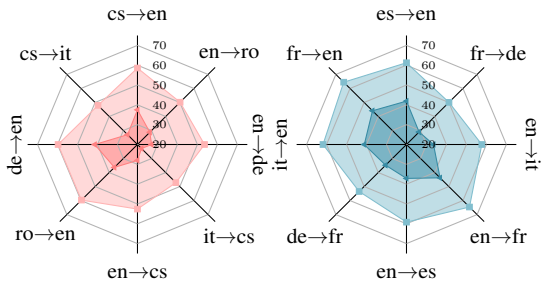

\subsection{Baselines}

We re-implemented \citet{reheman2023prompting}'s method which augments NMT systems via TM-based one-shot learning. For NMT systems, we chose two champion models in WMT: Facebook's WMT19 en $\leftrightarrow$ de models \cite{ng2019facebook} and WMT21 multilingual models \cite{tran2021facebook}. These WMT models were all trained on large-scale bilingual data and are improved by using a series of techniques, such as back-translation and fine-tuning. As a second baseline, we chose the \textit{k}NN-MT model \cite{khandelwal2020nearest} because it is a very strong model for TM and NMT combination.

\begin{table*}[t!]
\small
    \centering

 \resizebox{1.0\linewidth}{!}{
    \begin{tabular}{cc|cc|cc|cc|ccc}
        \toprule
        \multicolumn{2}{c|}{\multirow{3}{*}[-0.1ex]{\makecell{\textbf{Domain}}}} & \multicolumn{2}{c|}{\multirow{2}{*}[-1ex]{\normalsize{\textbf{$k$NN-MT}}}} & \multicolumn{2}{c|}{\normalsize{\textbf{WMT19 200M}}} & \multicolumn{2}{c|}{\normalsize{\textbf{WMT21 4B}}} & \multicolumn{3}{c}{\normalsize{\textbf{davinci-003 175B}}} \\
        & & & & \multicolumn{1}{c}{\multirow{2}{*}[-0.1ex]{\makecell{\textbf{NMT}}}} & \multicolumn{1}{c|}{\multirow{2}{*}[-0.1ex]{\makecell{\textbf{NMT+TM}}}} & \multicolumn{1}{c}{\multirow{2}{*}[-0.1ex]{\makecell{\textbf{NMT}}}} & \multicolumn{1}{c|}{\multirow{2}{*}[-0.1ex]{\makecell{\textbf{NMT+TM}}}} & \multicolumn{1}{c}{\multirow{2}{*}[0.4ex]{\textbf{LLM}}} &
        \multicolumn{1}{c}{\multirow{2}{*}[0.4ex]{\textbf{LLM+TM}}}  & \multicolumn{1}{c}{\multirow{2}{*}[0.4ex]{\textbf{LLM+TM}}}\\
        & & & & & & & & \multicolumn{1}{c}{\multirow{1}{*}[0.1ex]{\footnotesize{(zero-shot)}}} &\multicolumn{1}{c}{\multirow{1}{*}[0.1ex]{\footnotesize{(one-shot)}}} & \multicolumn{1}{c}{\multirow{1}{*}[0.1ex]{\footnotesize{(few-shot)}}} \\
        \midrule
        \multicolumn{2}{c|}{\multirow{1}{*}{IT}}   & \multicolumn{2}{c|}{45.82}&  38.09 & 40.63$_{(+2.54)}$ & 38.41 & \multicolumn{1}{l|}{46.61$_{(+8.20)}$}& 20.53 & 47.46$_{(+26.93)}$ & \textbf{51.03}$_{(+30.50)}$\\
        \multicolumn{2}{c|}{\multirow{1}{*}{Medical}}   & \multicolumn{2}{c|}{54.35}&  41.14 & 45.78$_{(+4.64)}$ & 47.94 & 55.36$_{(+7.42)}$ & 37.37 & 58.54$_{(+21.17)}$ & \textbf{60.40}$_{(+23.03)}$ \\
        \multicolumn{2}{c|}{\multirow{1}{*}{Koran}} &   \multicolumn{2}{c|}{19.45}&  17.11 & 17.53$_{(+0.42)}$ & \textbf{23.33} & 19.27$_{(-4.06)}$ & 17.59 & \multicolumn{1}{l}{18.80$_{(+1.21)}$} & \multicolumn{1}{l}{20.55$_{(+2.96)}$} \\
        \multicolumn{2}{c|}{\multirow{1}{*}{Law}}   & \multicolumn{2}{c|}{61.78}&  45.92 & 48.97$_{(+3.05)}$ & 51.60 & 59.97$_{(+8.37)}$ & 41.04 & 61.85$_{(+20.81)}$ & \textbf{64.92}$_{(+23.88)}$ \\

        \bottomrule
    \end{tabular}
    }

    \caption{Comparison of the NMT models and the $k$NN-MT model on the multi-domain dataset by BLEU. The COMET-22 version can be found in Table \ref{tab:app:multi_domains_by_COMET-22}.}
    \label{tab:t3}

\vspace{-0.5em}

\end{table*} 

\subsection{Translation Quality}

\paragraph{Main Results.} Table \ref{tab:t2} shows BLEU scores on the DGT-TM and JRC-A datasets. We see, first of all, that \textsc{TMPlm} achieves the best result among all the systems. When TMs are not involved, the performance of LLMs is 10 BLEU points lower than that of the NMT baselines. But, when armed with TMs, LLMs obtain very large BLEU improvements. The few-shot learning+LLM system even outperforms the strong NMT+TM baseline on all of the test sets. Also, by comparing the results of WMT19 200M models and WMT21 4B models, we see that larger models help more for making use of TM (see Section \ref{sec:s2} for more discussions). Besides, one-shot learning can give satisfactory results for \textsc{TMPlm} indicating that the most similar TM provides the most helpful translation hints. In Appendix \ref{app:few_shot_experimental} we will see that few-shot learning yields BLEU gains in a long-tail manner.

\begin{figure}[t!]
\centering
\begin{tikzpicture}
  \centering
    \scriptsize{
    \begin{axis}[
      at={(0,0)},
      ymajorgrids,
      grid style=dashed,
      legend style={at={(0.41,0.54)}, anchor=south west},
      legend cell align={left},
      ybar,
      enlarge x limits=0.5,
      xtick align=inside,
      height=.24\textwidth,
      width=.30\textwidth,
      bar width=0.8em,
      xlabel={(a) different models},
      symbolic x coords={w/o TM,w/ TM},
      legend style={cells={align=left}},
      xtick=data,
      nodes near coords align={vertical},
      ymin=18.00,
      ymax=69.00,
      ytick={29.00,50.00},
      xticklabels={w/o TM,w/ TM},
      ylabel style={yshift=-3em},xlabel style={yshift=0.3em,align=center},
      yticklabel style={/pgf/number format/fixed,/pgf/number format/fixed zerofill,/pgf/number format/precision=2,rotate=90},
      legend style={draw=none,
        line width=1pt,
        at={(0.5,1.0)},
        anchor=south},
        xtick=data,
      ]
         \addplot[fill=white,draw=red, area legend,postaction={pattern=north 
east lines},pattern color=red] coordinates {(w/o TM,20.91) (w/ TM,56.13) };
          \addplot[ fill=magenta!30,draw=red, area legend,postaction={pattern=crosshatch},pattern color=red] coordinates {(w/o TM,29.00) (w/ TM,57.39) };

          \addplot[fill=white, draw=blue, area legend,postaction={pattern=north 
east lines},pattern color=blue!80] coordinates {(w/o TM,30.31) (w/ TM,65.48) };
          \addplot[fill=cyan!30, draw=blue, area legend,postaction={pattern=crosshatch },pattern color=blue!80] coordinates {(w/o TM,38.89) (w/ TM,66.90) };
    \end{axis}
    } 
    \node [rectangle,draw=black,inner sep=2pt,minimum height=0.8em,minimum width=2.5em,font=\small,anchor=north,align=center,pattern=north east lines] (label1) at (1.5em,8.85em){};
    \node [rectangle,draw=black,inner sep=2pt,minimum height=0.8em,minimum width=2.5em,font=\small,anchor=north,align=center,pattern=crosshatch] (label2) at (1.5em,7.75em){};
    \node [align=center] (label1_1) at ([xshift=2.5em,yshift=-0.35em]label1.north){\tiny GPT3};
    \node [rectangle,draw=red,inner sep=2pt,minimum height=0.8em,minimum width=2.5em,font=\small,anchor=north,align=center] (a) at (4.5em,10.77em){};
    \node [align=center] (label1_2) at ([xshift=3.5em,yshift=-0.35em]label2.north){\tiny davinci-003};
    \scriptsize{
    \begin{axis}[
      at={(14.7em,0)},
      ymajorgrids,
      grid style=dashed,
      legend style={at={(0.02,0.65)}, anchor=south west},
      legend cell align={left},
      ybar,
      enlarge x limits=0.5,
      xtick align=inside,
      height=.24\textwidth,
      width=.30\textwidth,
      bar width=0.8em,
      xlabel={(b) different template styles},
      symbolic x coords={w/o TM,w/ TM},
      legend style={cells={align=left}},
      xtick=data,
      nodes near coords align={vertical},
      ymin=18.00,
      ymax=69.00,
      ytick={29.00,50},
      xticklabels={w/o TM,w/ TM},
      ylabel style={yshift=-3em},xlabel style={yshift=0.3em,align=center},
      yticklabel style={/pgf/number format/fixed,/pgf/number format/fixed zerofill,/pgf/number format/precision=2,rotate=90},
      legend style={draw=none,
        line width=1pt,
        at={(0.5,1.0)},
        anchor=south},
        xtick=data,
      ]
          \addplot[fill=white,draw=red, area legend,postaction={pattern=north 
east lines},pattern color=red] coordinates {(w/o TM,30.2) (w/ TM,56.88) };
          \addplot[ fill=magenta!30,draw=red, area legend,postaction={pattern=crosshatch},pattern color=red] coordinates {(w/o TM,29.00) (w/ TM,57.39) };
          \addplot[fill=white, draw=blue, area legend,postaction={pattern=north 
east lines},pattern color=blue!80] coordinates {(w/o TM,38.38) (w/ TM,63.97) };
          \addplot[fill=cyan!30, draw=blue, area legend,postaction={pattern=crosshatch },pattern color=blue!80] coordinates {(w/o TM,38.89) (w/ TM,66.90) };

    \end{axis}
    }
    \node [rectangle,draw=blue,inner sep=2pt,minimum height=0.8em,minimum width=2.5em,font=\small,anchor=north,align=center] (b) at (19.2em,10.77em){};
    
    \node [align=center] (b1) at ([xshift=3.25em,yshift=-0.35em]b.north){de $\rightarrow$ en};
    \node [rectangle,draw=black,inner sep=2pt,minimum height=1.25em,minimum width=27.8em,font=\small,anchor=north,align=center] (final) at (13.9em,11em){};

    \node [align=center] (a1) at ([xshift=3.25em,yshift=-0.35em]a.north){en $\rightarrow$ de};
    \node [rectangle,draw=black,inner sep=2pt,minimum height=0.8em,minimum width=2.5em,font=\small,anchor=north,align=center,pattern=north east lines] (label1) at (16.2em,8.85em){};
    \node [rectangle,draw=black,inner sep=2pt,minimum height=0.8em,minimum width=2.5em,font=\small,anchor=north,align=center,pattern=crosshatch] (label2) at (16.2em,7.75em){};
    \node [align=center] (label1_1) at ([xshift=3.25em,yshift=-0.35em]label1.north){\tiny instruction};
    \node [align=center] (label1_2) at ([xshift=2.25em,yshift=-0.35em]label2.north){\tiny code};
    \node [rotate=90]at (-1.7em,4.5em) {BLEU};

\end{tikzpicture}
\caption{Experiments on two impacts including different LLMs and different template styles.}
\label{fig:f3}
\vspace{-1.1em}
\end{figure}
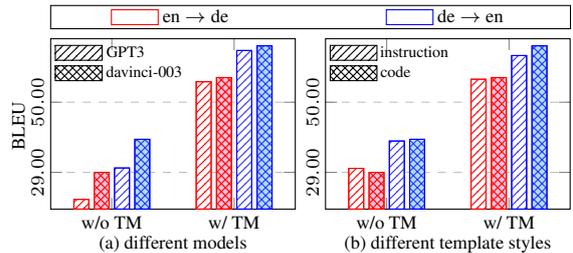

\paragraph{Multi-language Experiments.} We test \textsc{TMPlm} on more languages and run our system on data of 7 extra language pairs (i.e., 14 directions) from JRC-Acquis. From Figure \ref{fig:f2}, we see consistent improvements over all the language pairs. Even for non-English tasks, \textsc{TMPlm} can still achieve significant BLEU improvements. See Table \ref{tab:app.JRC_score} in Appendix \ref{app:Multi_languages_Complete} for complete experimental results.

\paragraph{Multi-domain Experiments.} Table \ref{tab:t3} shows BLEU results on the multi-domain dataset. Again, the \textsc{TMPlm} system is robust to the domain shift. It performs best on three of the four domains.

\begin{figure}
    \centering
    \definecolor{c1}{HTML}{81BECE}
\definecolor{c2}{HTML}{378BA4}
    \begin{tikzpicture}
    \tiny{
    \begin{axis}[
      at={(0,12em)},
      legend entries={de2en},
      ymajorgrids,
      xmajorgrids,
      grid style=dashed,
      xbar,
      legend image code/.code={%
                    \draw[#1, draw=none] (0cm,-0.1cm) rectangle (0.6cm,0.1cm);
                }, 
      height=.24\textwidth,
      width=.45\textwidth,
      bar width=1em,
      legend style={at={(21.2em,8.8em)}},
      symbolic y coords={ {few-shot\\(TM)}, {few-shot \\(out-domain)}, {few-shot\\ (in-domain)}, {zero-shot}},
      yticklabel style={align=right,font=\tiny},
      ytick=data,
      nodes near coords,
      nodes near coords align={horizontal},
      enlarge y limits=0.2,xticklabel style={/pgf/number format/fixed,/pgf/number format/fixed zerofill,/pgf/number format/precision=2},
      enlarge x limits=0.3,xticklabel style={/pgf/number format/fixed,/pgf/number format/fixed zerofill,/pgf/number format/precision=2},]
      \addplot[fill=red!30, draw=red] coordinates {(69.99,{few-shot\\(TM)}) (42.45,{few-shot \\(out-domain)}) (44.56,{few-shot\\ (in-domain)}) (38.89,{zero-shot})};
      \addlegendentry{de $\rightarrow$ en}
    \end{axis}
    \begin{axis}[
      at={(0em,0em)},
      ymajorgrids,
      xmajorgrids,
      grid style=dashed,
      xbar,
      height=.24\textwidth,
      width=.45\textwidth,
      bar width=1em,
      legend image code/.code={%
                    \draw[#1, draw=none] (0cm,-0.1cm) rectangle (0.6cm,0.1cm);
                }, 
      legend style={at={(21.2em,8.8em)}},
      symbolic y coords={ {few-shot\\(TM)}, {few-shot\\(out-domain)}, {few-shot\\(in-domain)}, {zero-shot}},
      yticklabel style={align=right,font=\tiny},
      ytick=data,
      nodes near coords,
      nodes near coords style={/pgf/number format/.cd,fixed zerofill,precision=2}, 
      nodes near coords align={horizontal},
      enlarge y limits=0.2,yticklabel style={/pgf/number format/fixed,/pgf/number format/fixed zerofill,/pgf/number format/precision=2},
      enlarge x limits=0.3,xticklabel style={/pgf/number format/fixed,/pgf/number format/fixed zerofill,/pgf/number format/precision=2},]
      \addplot[fill=c2!50, draw=c2] coordinates {(62.02,{few-shot\\(TM)}) (30.63,{few-shot\\(out-domain)}) (31.91,{few-shot\\(in-domain)}) (29.00,{zero-shot})};
      \addlegendentry{en $\rightarrow$ de}
    \end{axis}
  }
    \end{tikzpicture}
    \caption{BLEU scores of different prompting strategies on the DGT-TM dataset. In-domain and out-domain represent demonstrations randomly selected from the TM database of the DGT-TM dataset and \textit{newstest2017}, respectively. TM represents top-$k$ similar translation memories (i.e., demonstrations) retrieved from the TM database of the DGT-TM dataset.}
    \label{fig:f4}
\vspace{-1.1em}
\end{figure}
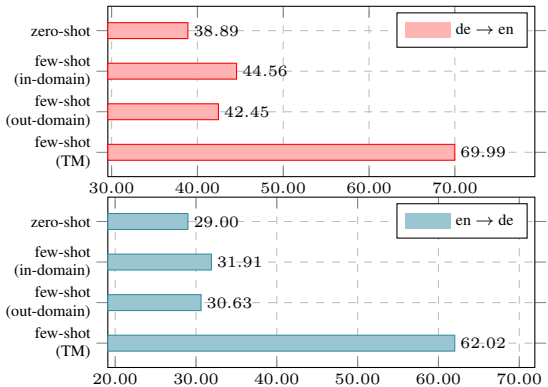

\subsection{Language Understanding Matters Most}
\label{sec:s2}

We then investigate an interesting issue: \textit{what kind of ability do large models have to make better use of TM-based prompts?} There are possibly three reasons, including the abilities of \textit{translating}, \textit{logically reasoning} and \textit{language understanding}. However, as seen from Table \ref{tab:t2}, the baseline LLMs are not strong translation systems and their BLEU scores are generally 10 points lower than the NMT systems. The translation ability of LLMs does not turn out to be important in TM-based prompting. Note that \texttt{davinci-003} is a successor of \texttt{GPT3} and is trained on additional large-scale code data. It has been pointed out that training LLMs on code data can lead to a strong ability of logical reasoning \cite{liang2022holistic}. As seen in Figure \ref{fig:f3} (a), however, no big difference between \texttt{davinci-003} and \texttt{GPT3} in BLEU performance. On the other hand, \texttt{davinci-003} has a significant ability to deal with instructions because it is tuned by using feedback to human instructions. Such a property makes \texttt{davinci-003} a better text processor, and thus a stronger translation system that works with various prompts. Therefore, it is the ability of language understanding that boosts LLMs' translation performance when prompted with TMs.

\subsection{Template Styles}

In Figure \ref{fig:f3} (b), we compare the performance between the code-style and instruction-style templates on the DGT-TM en-de and de-en tasks. For systems without TMs, the instruction-style template shows similar performance as the code-style template. However, when TMs are used, the code-style template is better in most cases. In Appendix \ref{app:dif_prompt_performance}, we test more templates and see a similar phenomenon that simpler templates work better.

\subsection{Prompting with randomly selected demonstrations}

We also compare the performance of \textsc{TMPlm} with the conventional few-shot, i.e., prompting LLM translators with randomly selected high quality demonstrations \cite{vilar2022prompting,DBLP:journals/corr/abs-2212-02437,zhang2023prompting,DBLP:journals/corr/abs-2301-13294,DBLP:journals/corr/abs-2302-09210}. We conduct experiments on the DGT-TM dataset, with demonstrations selected from the TM database of the DGT-TM dataset (in-domain) and \textit{newstest2017} (out-domain), respectively. In Figure \ref{fig:f4}, we see, \textsc{TMPlm} exceeds the conventional few-shot by about 30 BLEU points indicating that LLM can benefit from TMs much more than the conventional few-shot itself. It also demonstrates the valid information hinted by TMs as explained in Section \ref{sec:s1}.

\begin{figure}
    \centering
\tikzset{global scale/.style={
    scale=#1,
    every node/.append style={scale=#1}
  }
}

\pgfplotsset{
   width=0.4\textwidth,
   height=0.25\textheight,
   grid=major,
   major grid style={dotted},
   symbolic x coords={0,0.1,0.2,0.3,0.4,0.5,0.6,0.7,0.8,0.9,1.0},
   enlarge y limits={upper,value=0.05},
   legend style={
      fill,
      at={(0.50,16.5em)},
      legend columns=3,
      legend cell align=left,
      anchor=south
      },
   }
\begin{tikzpicture}
\useasboundingbox (13em,-1em) rectangle (3.5em,10em);
\begin{axis}[
global scale = 0.63,
   at={(0,1em)},
   xtick pos=left,
   axis y line*=right,
   bar width=0.2cm,
   ymin=0, ymax=1.000,
   ytick={0,0.25,0.50,0.75,1.00},
   xtick={0,0.1,0.2,0.3,0.4,0.5,0.6,0.7,0.8,0.9,1.0},
   xticklabels={0,,0.2,,0.4,,0.6,,0.8,,1.0},
   yticklabels={,,,,},
   ylabel style={align=center},
   xtick=data,
   xticklabel style={
      inner sep=0pt,
      anchor=north east,
      rotate=60
      },
    yticklabel style={/pgf/number format/fixed,/pgf/number format/fixed zerofill,/pgf/number format/precision=2,rotate=-90}
   ]              
   \addplot[very thick,draw=blue!90,mark=pentagon*,mark color=blue!30,color=blue!30] plot coordinates {
      (0,0) (0.1,0.0027) (0.2,0.0357)
      (0.3,0.1747) (0.4,0.299) (0.5,0.3747)
      (0.6,0.466) (0.7,0.5556) (0.8,0.6382)
      (0.9,0.7592) (1.0,1)
      };\label{B1}                 
\end{axis}
\begin{axis}[
global scale = 0.63,
   at={(0,1em)},
    legend entries={WMT19 200M},
    ymajorgrids=true,
    xmajorgrids=true,
    legend style={draw=none,
    line width=1pt,
    at={(5em,12.5em)},
    anchor=south},
   axis y line*=left,
   xticklabels={},
   ymin=20.00, ymax=70.00,
   ytick={20.00,32.50,45.00,57.5,70},
   yticklabel style={/pgf/number format/fixed,/pgf/number format/fixed zerofill,/pgf/number format/precision=2,rotate=90}
   ]
   \addplot[very thick,draw=red!20,mark=square*,mark color=red!20,color=red!20] plot coordinates{
      (0,66.90)(0.1,66.90) (0.2,66.88)
      (0.3,66.34) (0.4,65.63) (0.5,64.30)
      (0.6,62.70) (0.7,61.04) (0.8,58.45)
      (0.9,54.15) (1.0,44.21)
      };
   \addplot[very thick,draw=red!90,mark=triangle*,mark color=red!90,color=red!90] plot coordinates {
      (0,66.90) (0.1,66.90) (0.2,67.06)
      (0.3,67.02) (0.4,66.74) (0.5,65.84)
      (0.6,64.58) (0.7,63.08) (0.8,60.59)
      (0.9,57.33) (1.0,49.21)
      };
   \end{axis}
\node at (4em,-0.5em){\scriptsize (a) DGT-TM de $\rightarrow$ en};
\node at (13.2em,-0.5em){\scriptsize (b) IT domain de $\rightarrow$ en};
\begin{axis}[
global scale = 0.63,
   at={(9em,1em)},
   xtick pos=left,
   axis y line*=right,
   bar width=0.2cm,
   ymin=0, ymax=1.000,
   xticklabels={0,,0.2,,0.4,,0.6,,0.8,,1.0},
   ytick={0,0.250,0.5,0.750,1.000},
   yticklabels={0,0.25,0.50,0.75,1.00},
   ylabel style={align=center},
   ylabel style={rotate=180},
   xtick=data,
   xticklabel style={
      inner sep=0pt,
      anchor=north east,
      rotate=60
      },
    yticklabel style={/pgf/number format/fixed,/pgf/number format/fixed zerofill,/pgf/number format/precision=3,rotate=-90}
   ]   
   \addplot[very thick,draw=blue!90,mark=pentagon*,mark color=blue!30,color=blue!30] plot coordinates {
      (0,0) (0.1,0.062) (0.2,0.134)
      (0.3,0.2115) (0.4,0.314) (0.5,0.387)
      (0.6,0.598) (0.7,0.76) (0.8,0.8565)
      (0.9,0.9425) (1.0,1)
      };\label{B2}                
\end{axis}
\begin{axis}[
global scale = 0.63,
    at={(9em,1em)},
       legend entries={,WMT21 4B},
    ymajorgrids=true,
    xmajorgrids=true,
    legend style={draw=none,
    line width=1pt,
    at={(5em,12.5em)},
    anchor=south},
   axis y line*=left,
   xticklabels={},
   ymin=20.00, ymax=70.00,
   ytick={20.00,32.50,45.00,57.5,70},
   yticklabels={,,,,},
   yticklabel style={/pgf/number format/fixed,/pgf/number format/fixed zerofill,/pgf/number format/precision=2,rotate=90}
   ]
   \addplot[very thick,draw=red!20,mark=square*,mark color=red!20,color=red!20] plot coordinates{
      (0,47.46)(0.1,44.48) (0.2,43.85)
      (0.3,43.30) (0.4,40.78) (0.5,36.92)
      (0.6,35.78) (0.7,32.55) (0.8,29.27)
      (0.9,28.17) (1.0,26.25)
      };
   \addplot[very thick,draw=red!90,mark=triangle*,mark color=red!90,color=red!90] plot coordinates {
      (0,47.46) (0.1,44.01) (0.2,43.19)
      (0.3,42.31) (0.4,37.53) (0.5,34.35)
      (0.6,33.31) (0.7,30.17) (0.8,27.23)
      (0.9,26.25) (1.0,24.90)
      };
   \end{axis}
   \node [rectangle,draw=black,inner sep=2pt,minimum height=1em,minimum width=16.87em,font=\small,anchor=north,align=center] (final) at (8.4em,10.0em){};
   \node  [rotate=90] at(-1.3em,5em){\scriptsize BLEU};
   \node  [rotate=270] at(18.2em,5em){\scriptsize proportion};
\end{tikzpicture}
    \caption{BLEU scores as functions of thresholds of using similar sentences in TMs on the DGT-TM and IT domain data. The left $y$-axis represents the BLEU scores of prompting LLMs with the translation results from NMT systems, and the $x$-axis represents the similarity (i.e., the FMS in Appendix \ref{app: retrieve}) thresholds by which we have a trade-off between using TMs and NMT results as prompts (1 means that we only use TMs as prompts, and 0 means that we only use NMT outputs as prompts). Deep and light red curves represent the performance of the LLMs when working with the WMT19 200M and WMT21 4B systems. Blue curves represent the proportion of the use of TMs (see the right $y$-axis).}
    \label{fig:f5}
\vspace{-1.1em}
\end{figure}
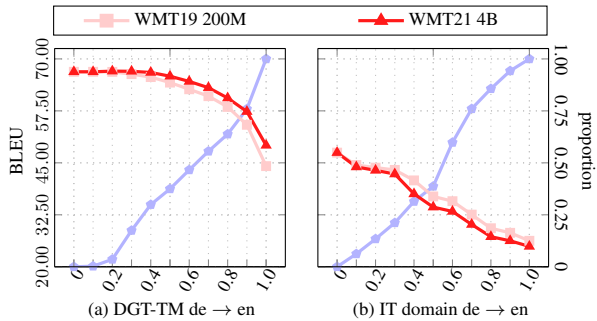

\subsection{Combining TMs and NMT results}

To examine the impact of high-quality translations on prompting LLMs, we replace the retrieved TM with the translation result of the WMT19 and WMT21 NMT systems when the TM's similarity is not high enough. We conducted experiments on the DGT-TM de $\rightarrow$ en data and the IT data in the multi-domain dataset because the sentence similarity distributes differently on them (see Appendix \ref{app: retrieve}). In Figure \ref{fig:f5}, we can see that the performance declines as more NMT translation results replace the TM results in prompting. This demonstrates that the quality of translations plays an important role in prompting LLMs. We also see that the performance on DGT-TM declines faster than that on IT domain. We attribute this to the better translation quality of the NMT models on the DGT-TM dataset.

There is an interesting finding that the method of prompting LLMs with the NMT results cannot surpass the NMT system itself, while the BLEU scores of prompting LLMs with TMs are always better than those of the TMs. It indicates that LLMs indeed process the prompting texts rather than simply outputting the prompting texts.

\section{Conclusion}

We have proposed \textsc{TMPlm}, an in-context learning method to prompt TMs for LLMs. By incorporating TMs into tailored templates, LLMs with \textsc{TMPlm} outperforms the state-of-the-art NMT models with TM prompting. We have also demonstrated that the ability of language understanding plays an important role in prompting LLMs with TMs. 

\section*{Limitations}

The similarity of TMs is an important factor influencing the translations of \textsc{TMPlm}. However, high-similarity TMs are not always available in practical applications. It is worth studying methods to make use of relatively low-similarity translations in LLM-based translation systems.

\section*{Acknowledgements}

This work was supported in part by the National Science Foundation of China (No. 62276056), the National Key R\&D Program of China, the China HTRD Center Project (No. 2020AAA0107904), the Natural Science Foundation of Liaoning Province of China (2022-KF-16-01), the Yunnan Provincial Major Science and Technology Special Plan Projects (No. 202103AA080015), the Fundamental Research Funds for the Central Universities (Nos. N2216016, N2216001, and N2216002), and the Program of Introducing Talents of Discipline to Universities, Plan 111 (No. B16009).

\bibliography{custom}

\begin{thebibliography}{28}
\expandafter\ifx\csname natexlab\endcsname\relax\def\natexlab#1{#1}\fi

\bibitem[{Agrawal et~al.(2022)Agrawal, Zhou, Lewis, Zettlemoyer, and
  Ghazvininejad}]{DBLP:journals/corr/abs-2212-02437}
Sweta Agrawal, Chunting Zhou, Mike Lewis, Luke Zettlemoyer, and Marjan
  Ghazvininejad. 2022.
\newblock \href {https://doi.org/10.48550/arXiv.2212.02437} {In-context
  examples selection for machine translation}.
\newblock \emph{CoRR}, abs/2212.02437.

\bibitem[{Aharoni and Goldberg(2020)}]{aharoni2020unsupervised}
Roee Aharoni and Yoav Goldberg. 2020.
\newblock \href {https://doi.org/10.18653/v1/2020.acl-main.692} {Unsupervised
  domain clusters in pretrained language models}.
\newblock In \emph{Proceedings of the 58th Annual Meeting of the Association
  for Computational Linguistics, {ACL} 2020, Online, July 5-10, 2020}, pages
  7747--7763. Association for Computational Linguistics.

\bibitem[{Bialecki et~al.(2012)Bialecki, Muir, and
  Ingersoll}]{bialecki2012apache}
Andrzej Bialecki, Robert Muir, and Grant Ingersoll. 2012.
\newblock Apache lucene 4.
\newblock In \emph{Proceedings of the {SIGIR} 2012 Workshop on Open Source
  Information Retrieval, OSIR@SIGIR 2012, Portland, Oregon, USA, 16th August
  2012}, pages 17--24. University of Otago, Dunedin, New Zealand.

\bibitem[{Bult{\'{e}} and Tezcan(2019)}]{bulte2019neural}
Bram Bult{\'{e}} and Arda Tezcan. 2019.
\newblock \href {https://doi.org/10.18653/v1/p19-1175} {Neural fuzzy repair:
  Integrating fuzzy matches into neural machine translation}.
\newblock In \emph{Proceedings of the 57th Conference of the Association for
  Computational Linguistics, {ACL} 2019, Florence, Italy, July 28- August 2,
  2019, Volume 1: Long Papers}, pages 1800--1809. Association for Computational
  Linguistics.

\bibitem[{Freitag et~al.(2022)Freitag, Rei, Mathur, Lo, Stewart, Avramidis,
  Kocmi, Foster, Lavie, and Martins}]{DBLP:conf/wmt/FreitagRMLSAKFLM22}
Markus Freitag, Ricardo Rei, Nitika Mathur, Chi{-}kiu Lo, Craig Stewart,
  Eleftherios Avramidis, Tom Kocmi, George~F. Foster, Alon Lavie, and
  Andr{\'{e}} F.~T. Martins. 2022.
\newblock \href {https://aclanthology.org/2022.wmt-1.2} {Results of {WMT22}
  metrics shared task: Stop using {BLEU} - neural metrics are better and more
  robust}.
\newblock In \emph{Proceedings of the Seventh Conference on Machine
  Translation, {WMT} 2022, Abu Dhabi, United Arab Emirates (Hybrid), December
  7-8, 2022}, pages 46--68. Association for Computational Linguistics.

\bibitem[{Gu et~al.(2018)Gu, Wang, Cho, and Li}]{gu2018search}
Jiatao Gu, Yong Wang, Kyunghyun Cho, and Victor O.~K. Li. 2018.
\newblock \href
  {https://www.aaai.org/ocs/index.php/AAAI/AAAI18/paper/view/17282} {Search
  engine guided neural machine translation}.
\newblock In \emph{Proceedings of the Thirty-Second {AAAI} Conference on
  Artificial Intelligence, (AAAI-18), the 30th innovative Applications of
  Artificial Intelligence (IAAI-18), and the 8th {AAAI} Symposium on
  Educational Advances in Artificial Intelligence (EAAI-18), New Orleans,
  Louisiana, USA, February 2-7, 2018}, pages 5133--5140. {AAAI} Press.

\bibitem[{He et~al.(2021)He, Huang, Cui, Li, and Liu}]{he2021fast}
Qiuxiang He, Guoping Huang, Qu~Cui, Li~Li, and Lemao Liu. 2021.
\newblock \href {https://doi.org/10.18653/v1/2021.acl-long.246} {Fast and
  accurate neural machine translation with translation memory}.
\newblock In \emph{Proceedings of the 59th Annual Meeting of the Association
  for Computational Linguistics and the 11th International Joint Conference on
  Natural Language Processing, {ACL/IJCNLP} 2021, (Volume 1: Long Papers),
  Virtual Event, August 1-6, 2021}, pages 3170--3180. Association for
  Computational Linguistics.

\bibitem[{Hendy et~al.(2023)Hendy, Abdelrehim, Sharaf, Raunak, Gabr,
  Matsushita, Kim, Afify, and Awadalla}]{DBLP:journals/corr/abs-2302-09210}
Amr Hendy, Mohamed Abdelrehim, Amr Sharaf, Vikas Raunak, Mohamed Gabr, Hitokazu
  Matsushita, Young~Jin Kim, Mohamed Afify, and Hany~Hassan Awadalla. 2023.
\newblock \href {https://doi.org/10.48550/arXiv.2302.09210} {How good are {GPT}
  models at machine translation? {A} comprehensive evaluation}.
\newblock \emph{CoRR}, abs/2302.09210.

\bibitem[{Hossain et~al.(2020)Hossain, Ghazvininejad, and
  Zettlemoyer}]{hossain2020simple}
Nabil Hossain, Marjan Ghazvininejad, and Luke Zettlemoyer. 2020.
\newblock \href {https://doi.org/10.18653/v1/2020.acl-main.228} {Simple and
  effective retrieve-edit-rerank text generation}.
\newblock In \emph{Proceedings of the 58th Annual Meeting of the Association
  for Computational Linguistics, {ACL} 2020, Online, July 5-10, 2020}, pages
  2532--2538. Association for Computational Linguistics.

\bibitem[{Khandelwal et~al.(2021)Khandelwal, Fan, Jurafsky, Zettlemoyer, and
  Lewis}]{khandelwal2020nearest}
Urvashi Khandelwal, Angela Fan, Dan Jurafsky, Luke Zettlemoyer, and Mike Lewis.
  2021.
\newblock \href {https://openreview.net/forum?id=7wCBOfJ8hJM} {Nearest neighbor
  machine translation}.
\newblock In \emph{9th International Conference on Learning Representations,
  {ICLR} 2021, Virtual Event, Austria, May 3-7, 2021}. OpenReview.net.

\bibitem[{Liang et~al.(2022)Liang, Bommasani, Lee, Tsipras, Soylu, Yasunaga,
  Zhang, Narayanan, Wu, Kumar et~al.}]{liang2022holistic}
Percy Liang, Rishi Bommasani, Tony Lee, Dimitris Tsipras, Dilara Soylu,
  Michihiro Yasunaga, Yian Zhang, Deepak Narayanan, Yuhuai Wu, Ananya Kumar,
  et~al. 2022.
\newblock Holistic evaluation of language models.
\newblock \emph{arXiv preprint arXiv:2211.09110}.

\bibitem[{Meng et~al.(2022)Meng, Li, Zheng, Wu, Sun, Zhang, and
  Li}]{meng2021fast}
Yuxian Meng, Xiaoya Li, Xiayu Zheng, Fei Wu, Xiaofei Sun, Tianwei Zhang, and
  Jiwei Li. 2022.
\newblock \href {https://doi.org/10.18653/v1/2022.findings-acl.47} {Fast
  nearest neighbor machine translation}.
\newblock In \emph{Findings of the Association for Computational Linguistics:
  {ACL} 2022, Dublin, Ireland, May 22-27, 2022}, pages 555--565. Association
  for Computational Linguistics.

\bibitem[{Moslem et~al.(2023)Moslem, Haque, and
  Way}]{DBLP:journals/corr/abs-2301-13294}
Yasmin Moslem, Rejwanul Haque, and Andy Way. 2023.
\newblock \href {https://doi.org/10.48550/arXiv.2301.13294} {Adaptive machine
  translation with large language models}.
\newblock \emph{CoRR}, abs/2301.13294.

\bibitem[{Ng et~al.(2019)Ng, Yee, Baevski, Ott, Auli, and
  Edunov}]{ng2019facebook}
Nathan Ng, Kyra Yee, Alexei Baevski, Myle Ott, Michael Auli, and Sergey Edunov.
  2019.
\newblock \href {https://doi.org/10.18653/v1/w19-5333} {Facebook fair's {WMT19}
  news translation task submission}.
\newblock In \emph{Proceedings of the Fourth Conference on Machine Translation,
  {WMT} 2019, Florence, Italy, August 1-2, 2019 - Volume 2: Shared Task Papers,
  Day 1}, pages 314--319. Association for Computational Linguistics.

\bibitem[{Ouyang et~al.(2022)Ouyang, Wu, Jiang, Almeida, Wainwright, Mishkin,
  Zhang, Agarwal, Slama, Ray et~al.}]{ouyang2022training}
Long Ouyang, Jeff Wu, Xu~Jiang, Diogo Almeida, Carroll~L Wainwright, Pamela
  Mishkin, Chong Zhang, Sandhini Agarwal, Katarina Slama, Alex Ray, et~al.
  2022.
\newblock Training language models to follow instructions with human feedback.
\newblock \emph{arXiv preprint arXiv:2203.02155}.

\bibitem[{Reheman et~al.(2023)Reheman, Zhou, Luo, Yang, Xiao, and
  Zhu}]{reheman2023prompting}
Abudurexiti Reheman, Tao Zhou, Yingfeng Luo, Di~Yang, Tong Xiao, and Jingbo
  Zhu. 2023.
\newblock Prompting neural machine translation with translation memories.
\newblock \emph{arXiv preprint arXiv:2301.05380}.

\bibitem[{Rei et~al.(2022)Rei, de~Souza, Alves, Zerva, Farinha, Glushkova,
  Lavie, Coheur, and Martins}]{DBLP:conf/wmt/ReiSAZFGLCM22}
Ricardo Rei, Jos{\'{e}} G.~C. de~Souza, Duarte~M. Alves, Chrysoula Zerva,
  Ana~C. Farinha, Taisiya Glushkova, Alon Lavie, Lu{\'{\i}}sa Coheur, and
  Andr{\'{e}} F.~T. Martins. 2022.
\newblock \href {https://aclanthology.org/2022.wmt-1.52} {{COMET-22:}
  unbabel-ist 2022 submission for the metrics shared task}.
\newblock In \emph{Proceedings of the Seventh Conference on Machine
  Translation, {WMT} 2022, Abu Dhabi, United Arab Emirates (Hybrid), December
  7-8, 2022}, pages 578--585. Association for Computational Linguistics.

\bibitem[{Steinberger et~al.(2012)Steinberger, Eisele, Klocek, Pilos, and
  Schl{\"{u}}ter}]{Ralf2012dgt}
Ralf Steinberger, Andreas Eisele, Szymon Klocek, Spyridon Pilos, and Patrick
  Schl{\"{u}}ter. 2012.
\newblock \href
  {http://www.lrec-conf.org/proceedings/lrec2012/summaries/814.html} {{DGT-TM:}
  {A} freely available translation memory in 22 languages}.
\newblock In \emph{Proceedings of the Eighth International Conference on
  Language Resources and Evaluation, {LREC} 2012, Istanbul, Turkey, May 23-25,
  2012}, pages 454--459. European Language Resources Association {(ELRA)}.

\bibitem[{Steinberger et~al.(2006)Steinberger, Pouliquen, Widiger, Ignat,
  Erjavec, Tufis, and Varga}]{steinberger2006jrc}
Ralf Steinberger, Bruno Pouliquen, Anna Widiger, Camelia Ignat, Tomaz Erjavec,
  Dan Tufis, and D{\'{a}}niel Varga. 2006.
\newblock \href
  {http://www.lrec-conf.org/proceedings/lrec2006/summaries/340.html} {The
  jrc-acquis: {A} multilingual aligned parallel corpus with 20+ languages}.
\newblock In \emph{Proceedings of the Fifth International Conference on
  Language Resources and Evaluation, {LREC} 2006, Genoa, Italy, May 22-28,
  2006}, pages 2142--2147. European Language Resources Association {(ELRA)}.

\bibitem[{Thoppilan et~al.(2022)Thoppilan, De~Freitas, Hall, Shazeer,
  Kulshreshtha, Cheng, Jin, Bos, Baker, Du et~al.}]{thoppilan2022lamda}
Romal Thoppilan, Daniel De~Freitas, Jamie Hall, Noam Shazeer, Apoorv
  Kulshreshtha, Heng-Tze Cheng, Alicia Jin, Taylor Bos, Leslie Baker, Yu~Du,
  et~al. 2022.
\newblock Lamda: Language models for dialog applications.
\newblock \emph{arXiv preprint arXiv:2201.08239}.

\bibitem[{Tiedemann(2012)}]{tiedemann2012parallel}
J{\"{o}}rg Tiedemann. 2012.
\newblock \href
  {http://www.lrec-conf.org/proceedings/lrec2012/summaries/463.html} {Parallel
  data, tools and interfaces in {OPUS}}.
\newblock In \emph{Proceedings of the Eighth International Conference on
  Language Resources and Evaluation, {LREC} 2012, Istanbul, Turkey, May 23-25,
  2012}, pages 2214--2218. European Language Resources Association {(ELRA)}.

\bibitem[{Tran et~al.(2021)Tran, Bhosale, Cross, Koehn, Edunov, and
  Fan}]{tran2021facebook}
Chau Tran, Shruti Bhosale, James Cross, Philipp Koehn, Sergey Edunov, and
  Angela Fan. 2021.
\newblock \href {https://aclanthology.org/2021.wmt-1.19} {Facebook ai's {WMT21}
  news translation task submission}.
\newblock In \emph{Proceedings of the Sixth Conference on Machine Translation,
  WMT@EMNLP 2021, Online Event, November 10-11, 2021}, pages 205--215.
  Association for Computational Linguistics.

\bibitem[{Vilar et~al.(2022)Vilar, Freitag, Cherry, Luo, Ratnakar, and
  Foster}]{vilar2022prompting}
David Vilar, Markus Freitag, Colin Cherry, Jiaming Luo, Viresh Ratnakar, and
  George Foster. 2022.
\newblock Prompting palm for translation: Assessing strategies and performance.
\newblock \emph{arXiv preprint arXiv:2211.09102}.

\bibitem[{Wei et~al.(2022)Wei, Tay, Bommasani, Raffel, Zoph, Borgeaud,
  Yogatama, Bosma, Zhou, Metzler et~al.}]{wei2022emergent}
Jason Wei, Yi~Tay, Rishi Bommasani, Colin Raffel, Barret Zoph, Sebastian
  Borgeaud, Dani Yogatama, Maarten Bosma, Denny Zhou, Donald Metzler, et~al.
  2022.
\newblock Emergent abilities of large language models.
\newblock \emph{arXiv preprint arXiv:2206.07682}.

\bibitem[{Xiao et~al.(2012)Xiao, Zhu, Zhang, and Li}]{xiao2012niutrans}
Tong Xiao, Jingbo Zhu, Hao Zhang, and Qiang Li. 2012.
\newblock \href {https://aclanthology.org/P12-3004/} {Niutrans: An open source
  toolkit for phrase-based and syntax-based machine translation}.
\newblock In \emph{The 50th Annual Meeting of the Association for Computational
  Linguistics, Proceedings of the System Demonstrations, July 10, 2012, Jeju
  Island, Korea}, pages 19--24. The Association for Computer Linguistics.

\bibitem[{Xu et~al.(2020)Xu, Crego, and Senellart}]{xu2020boosting}
Jitao Xu, Josep~Maria Crego, and Jean Senellart. 2020.
\newblock \href {https://doi.org/10.18653/v1/2020.acl-main.144} {Boosting
  neural machine translation with similar translations}.
\newblock In \emph{Proceedings of the 58th Annual Meeting of the Association
  for Computational Linguistics, {ACL} 2020, Online, July 5-10, 2020}, pages
  1580--1590. Association for Computational Linguistics.

\bibitem[{Zhang et~al.(2023)Zhang, Haddow, and Birch}]{zhang2023prompting}
Biao Zhang, Barry Haddow, and Alexandra Birch. 2023.
\newblock Prompting large language model for machine translation: A case study.
\newblock \emph{arXiv preprint arXiv:2301.07069}.

\bibitem[{Zhang et~al.(2018)Zhang, Utiyama, Sumita, Neubig, and
  Nakamura}]{zhang2018guiding}
Jingyi Zhang, Masao Utiyama, Eiichiro Sumita, Graham Neubig, and Satoshi
  Nakamura. 2018.
\newblock \href {https://doi.org/10.18653/v1/n18-1120} {Guiding neural machine
  translation with retrieved translation pieces}.
\newblock In \emph{Proceedings of the 2018 Conference of the North American
  Chapter of the Association for Computational Linguistics: Human Language
  Technologies, {NAACL-HLT} 2018, New Orleans, Louisiana, USA, June 1-6, 2018,
  Volume 1 (Long Papers)}, pages 1325--1335. Association for Computational
  Linguistics.

\end{thebibliography}
\bibliographystyle{acl_natbib}

\appendix

\section{Detailed Experimental Setup}
\label{app:appendix}

\begin{table*}[t!]
\small
\centering
\begin{tabular}{c|c|c|c|ccccc}
\toprule
\multirow{2}{*}{\textbf{Dataset}} & \multirow{2}{*}{\textbf{Lang}} & \multirow{2}{*}{\textbf{Domain}} & \multirow{2}{*}{\textbf{TM scale}} & \multicolumn{5}{c}{\textbf{FMS}} \\ 
& & & & [0, 0.2) & [0.2, 0.4) & [0.4, 0.6) & [0.6, 0.8) & [0.8, 1.0) \\
\midrule
\multirow{2}{*}{DGT-TM}& En-De & - & 3.1M & 2\% & 23\% & 16\% & 17\% & 42\% \\
& De-En & - & 3.1M & 4\% & 26\% & 17\% & 17\% & 36\% \\
\midrule
\multirow{16}{*}{JRC-A}& En-De & - & 423K & 6\% & 33\% & 18\% & 13\% & 30\% \\
& De-En & - & 423K & 6\% & 33\% & 18\% & 15\% & 28\% \\
& En-Fr & - & 424K & 3\% & 34\% & 19\% & 14\% & 30\% \\
& Fr-En & - & 424K & 3\% & 33\% & 19\% & 15\% & 30\% \\
& De-Fr & - & 846K & 9\% & 34\% & 16\% & 12\% & 29\% \\
& Fr-De & - & 846K & 8\% & 34\% & 16\% & 12\% & 30\% \\
& En-It & - & 433K & 7\% & 32\% & 18\% & 14\% & 29\% \\
& It-En & - & 433K & 7\% & 32\% & 17\% & 16\% & 28\% \\
& En-Ro & - & 273K & 7\% & 39\% & 21\% & 14\% & 19\% \\
& Ro-En & - & 273K & 6\% & 37\% & 22\% & 15\% & 20\% \\
& En-Es & - & 432K & 2\% & 34\% & 20\% & 16\% & 28\% \\
& Es-En & - & 432K & 2\% & 34\% & 20\% & 16\% & 28\% \\
& En-Cs & - & 681K & 12\% & 33\% & 17\% & 12\% & 26\% \\
& Cs-En & - & 681K & 13\% & 32\% & 15\% & 13\% & 27\% \\
& Cs-It & - & 714K & 11\% & 31\% & 17\% & 14\% & 27\% \\
& It-En & - & 714K & 12\% & 32\% & 16\% & 13\% & 27\% \\
\midrule
\multirow{4}{*}{multi-domain}& De-En & IT & 223K & 14\% & 18\% & 28\% & 26\% & 14\% \\
& De-En & Koran & 18K & 2\% & 26\% & 33\% & 28\% & 11\% \\
& De-En & Law & 467K & 8\% & 31\% & 18\% & 14\% & 28\% \\
& De-En & Medical & 248K & 7\% & 23\% & 20\% & 17\% & 33\% \\
\midrule
WMT14 & En-De & - & 4.5M & 18\% & 68\% & 12\% & 1\% & 1\% \\
\midrule
WMT19 & De-En & - & 30M & 13\% & 65\% & 19\% & 2\% & 1\%\\
\bottomrule
\end{tabular}
\caption{TMs and proportions of the retrieved sentences in different ranges of FMS.}
 \label{tab:app.data_introduction}
\end{table*}

\subsection{Retrieval of Similar Sentences}
\label{app: retrieve}
Following \citealp{reheman2023prompting}, we adopt a word-level fuzzy matching strategy, with the numbers and punctuation marks removed. Specifically, we first use the search engine Apache Lucene \cite{bialecki2012apache} to acquire the Top$500$ similar TMs from TM database, then rerank the most similar TM by using the length normalized Levenshtein Distance, given by
\begin{eqnarray}
\mathrm{FMS}(X,S\mathrm)=1-\frac{\mathrm{LD}(X,S)}{\textrm{max}(|X|,|S|)}
\end{eqnarray}
where $\mathrm{FMS}(\cdot,\cdot)$ denotes the Fuzzy Match Score, $\mathrm{LD}(\cdot,\cdot)$ denotes the word level Levenshtein Distance, and $|\cdot|$ denotes the length of a sentence.

\subsection{Details of Datasets}
Datasets and their language directions used in our experiments are listed here.
\begin{itemize}
\item The DGT-TM dataset\cite{tiedemann2012parallel}, which is bidirectional in English-German; 
\item The JRC-Acquis (JRC-A) dataset\cite{steinberger2006jrc}, which includes 8 language pairs and 16 directions: English-German, English-French, German-French, English-Italian, English-Romanian, English-Spanish, English-Czech, and Czech-Italian;
\item The multi-domain dataset \cite{aharoni2020unsupervised}, which includes 4 domains in the German to English direction: Medical, Law, IT, and Koran.
 \end{itemize}

The statistics of these TM and the corresponding similarity ratios of retrieved sentences in the FMS metric are shown in Table \ref{tab:app.data_introduction}.

\subsection{Data Pre-processing}
\label{app:data pre-process}
For the DGT-TM, JRC-A and multi-domain datasets, we clean the data using the scripts provided by \citet{reheman2023prompting}'s work. To construct the test set and TM database for the DGT-TM and JRC-A datasets, we process each language direction separately. Specifically, we randomly extract 3,000 sentence pairs from each dataset as the test set, and use the remaining sentence pairs as the TM database. For the multi-domain dataset, we use its original test set as our test set and its original training set as the TM database. We use the FMS algorithm on the split data to obtain the TM corresponding to the test set. In particular, for the few-shot experiments, we retrieved the $k$ most similar sentence pairs from the TM database for each test sentence.

Finally, we replace the escaped characters in the dataset and use Moses\footnote{\url{http://www.statmt.org/moses/}} decoder detokenizer to recover the tokenized data before feeding it to the \texttt{davinci-003} system.

\subsection{Data Post-processing}
\label{app:data post-process}
\texttt{davinci-003} always generates redundant symbols at the beginning and end of sentences, including: `"', `$\backslash$n', `[', `]', and other escaped characters. The occurrences of these characters is regular and can be removed uniformly by scripts. Consequently, before scoring, we use NiuTrans \cite{xiao2012niutrans} word segmentation tool for Chinese and
Moses decoder's \texttt{tokenizer.perl} for all other languages. Finally we use \texttt{multi-bleu.perl} for scoring.

\subsection{More Prompt Templates}
We try a large number of prompt templates, as shown in Table \ref{tab:app.diff_prompt}. Without special specification, the instruction-style template with TM is the \#1, and without TM is the \#2, and the code-style template with TM is the \#17, and without TM is the \#18. In particular, in the multi-language experiment, we use the instruction-style template. The template for all of the few-shot experiments is obtained by increasing the number of TMs in \#17.

Punctuation has a significant impact on the generation results. For example, using template \#13, if the source sentence ends with `:', it will lead the model to continue generating words but not stop in an appropriate number of decoding steps. Meanwhile, although many templates have a similar form, their performance still differs. We believe that adding a strong boundary signal to the templates helps the model to know where to end.

\begin{table*}[t!]
\small
\centering
\scalebox{0.85}{
\begin{tabular}{c|lclc}
\toprule
\multirow{2}{*}{\textbf{No.}} & \multicolumn{1}{c}{\multirow{2}{*}{\textbf{Prompt Template}}} & \textbf{With} & \multicolumn{1}{c}{\multirow{2}{*}{\textbf{Sample}}} & \multirow{2}{*}{\textbf{BLEU}}\\ 
 & & \textbf{TM} & & \\
\midrule

\multirow{4}{*}{1} & If the translation of "$\mathbf{x}_{\mathrm{tm}}$" from \textit{src-lang} & \multirow{4}{*}{Yes} &  If the translation of "I have an apple." from English & \multirow{4}{*}{63.97} \\ 
& to \textit{tgt-lang} is "$\mathbf{y}_{\mathrm{tm}}$", then what is the &  & to German is "Ich habe einen Apfel." then what is the & \\
& translation of "x" from \textit{src-lang} to \textit{tgt-lang}? &  & translation of "I have an orange." from English to German? & \\
& Only translation results are required. &  & Only translation results are required. & \\

\midrule

\multirow{3}{*}{2} & What is the translation of "x" & \multirow{3}{*}{No} &  What is the translation of "I have an apple." & \multirow{3}{*}{38.38}\\ 
& from \textit{src-lang} to \textit{tgt-lang}? &  & from English to German? &\\
& Only translation results are required. &  & Only translation results are required. & \\

\midrule

\multirow{4}{*}{3} & If "$\mathbf{x}_{\mathrm{tm}}$" translated into \textit{tgt-lang} is "$\mathbf{y}_{\mathrm{tm}}$", & \multirow{4}{*}{Yes} &  If "I have an apple." translated into German & \multirow{4}{*}{61.9}\\ 
& then what is the translation of "x" &  & is "Ich habe einen Apfel." then what is the &\\
& should be if translated into \textit{tgt-lang}? &  & translation of "I have an orange." should be if translated &\\
& Only translation results are required. &  & into German? Only translation results are required. &\\

\midrule

\multirow{3}{*}{4} & What is the translation of "x" & \multirow{3}{*}{No} &  What is the translation of "I have an apple." & \multirow{3}{*}{37.93}\\ 
& should be if translated into \textit{tgt-lang}? &  & should be if translated into German? &\\
& Only translation results are required. &  & Only translation results are required. &\\

\midrule

\multirow{4}{*}{5} & If [$\mathbf{x}_{\mathrm{tm}}$] translated into \textit{tgt-lang} is [$\mathbf{y}_{\mathrm{tm}}$], & \multirow{4}{*}{Yes} &  If [I have an apple.] translated into German & \multirow{4}{*}{61.78}\\ 
& then what is the translation of [x] &  & is [Ich habe einen Apfel.] then what is the &\\
& should be if translated into \textit{tgt-lang}? &  & translation of [I have an orange.] should be if translated &\\
& Only translation results are required. &  & into German? Only translation results are required. &\\

\midrule

\multirow{5}{*}{6} & Translate \textit{src-lang} to \textit{tgt-lang}.\textbackslash n & \multirow{5}{*}{Yes} & Translate English to German.\textbackslash n& \multirow{5}{*}{65.25} \\ 
& [\textit{src-lang}]: [$\mathbf{x}_{\mathrm{tm}}$]\textbackslash n &  & [English]: [I have an apple.]\textbackslash n&\\
& [\textit{tgt-lang}]: [$\mathbf{y}_{\mathrm{tm}}$]\textbackslash n &  & [German]: [Ich habe einen Apfel.]\textbackslash n&\\
& [\textit{src-lang}]: [x]\textbackslash n &  & [English]: [I have an orange.]\textbackslash n&\\
& [\textit{tgt-lang}]: &  & [German]:&\\

\midrule

\multirow{5}{*}{7} & Translate \textit{src-lang} to \textit{tgt-lang}.\textbackslash n & \multirow{5}{*}{Yes} & Translate English to German.\textbackslash n& \multirow{5}{*}{66.02} \\ 
& [\textit{src-lang}]=[$\mathbf{x}_{\mathrm{tm}}$]\textbackslash n &  & [English]=[I have an apple.]\textbackslash n&\\
& [\textit{tgt-lang}]=[$\mathbf{y}_{\mathrm{tm}}$]\textbackslash n &  & [German]=[Ich habe einen Apfel.]\textbackslash n&\\
& [\textit{src-lang}]=[x]\textbackslash n &  & [English]=[I have an orange.]\textbackslash n&\\
& [\textit{tgt-lang}]= &  & [German]= &\\

\midrule

\multirow{3}{*}{8} & Translate \textit{src-lang} to \textit{tgt-lang}. & \multirow{3}{*}{Yes} & Translate English to German. & \multirow{3}{*}{66.08}\\ 
& [\textit{src-lang}]=[$\mathbf{x}_{\mathrm{tm}}$] [\textit{tgt-lang}]=[$\mathbf{y}_{\mathrm{tm}}$] &  & [English]=[I have an apple.] [German]=[Ich habe einen Apfel.] &\\
& [\textit{src-lang}]=[x] [\textit{tgt-lang}]= & & [English]=[I have an orange.] [German]=&\\

\midrule

\multirow{3}{*}{9} & Translate \textit{src-lang} to \textit{tgt-lang}.\textbackslash n & \multirow{3}{*}{Yes} & Translate English to German.\textbackslash n& \multirow{3}{*}{66.20}\\ 
& [\textit{src-lang}]=[$\mathbf{x}_{\mathrm{tm}}$] [\textit{tgt-lang}]=[$\mathbf{y}_{\mathrm{tm}}$]\textbackslash n &  & [English]=[I have an apple.] [German]=[Ich habe einen Apfel.]\textbackslash n&\\
& [\textit{src-lang}]=[x] [\textit{tgt-lang}]= &  & [English]=[I have an orange.] [German]= &\\

\midrule

\multirow{2}{*}{10} & if \textit{src-lang} = [$\mathbf{x}_{\mathrm{tm}}$] then \textit{tgt-lang} = [$\mathbf{y}_{\mathrm{tm}}$]; & \multirow{2}{*}{Yes} &  if English = [I have an apple.] then German = [Ich habe einen Apfel.]; & \multirow{2}{*}{66.75}\\ 
& if \textit{src-lang} = [x] then \textit{tgt-lang} = &  & if English = [I have an orange.] then German = &\\

\midrule

\multirow{2}{*}{11} & \textit{src-lang}="$\mathbf{x}_{\mathrm{tm}}$" \textit{tgt-lang}="$\mathbf{y}_{\mathrm{tm}}$" & \multirow{2}{*}{Yes} & English="I have an apple." German="Ich habe einen Apfel." & \multirow{2}{*}{66.28}\\ 
& \textit{src-lang}="x" \textit{tgt-lang}= &  & English="I have an orange." German= &\\

\midrule

\multirow{2}{*}{12} & \textit{src-lang}=[$\mathbf{x}_{\mathrm{tm}}$] \textit{tgt-lang}=[$\mathbf{y}_{\mathrm{tm}}$] & \multirow{2}{*}{Yes} & English=[I have an apple.] German=[Ich habe einen Apfel.] & \multirow{2}{*}{65.37}\\ 
& \textit{src-lang}=[x] \textit{tgt-lang}= &  & English=[I have an orange.] German= &\\

\midrule

\multirow{2}{*}{13} & [\textit{src-lang}] $\mathbf{x}_{\mathrm{tm}}$ [\textit{tgt-lang}] $\mathbf{y}_{\mathrm{tm}}$ & \multirow{2}{*}{Yes} & [English] I have an apple. [German] Ich habe einen Apfel.  & \multirow{2}{*}{58.47}\\ 
& [\textit{src-lang}] x [\textit{tgt-lang}] &  & [English] I have an orange. [German] &\\

\midrule

\multirow{2}{*}{14} & [\textit{src-lang}]: [$\mathbf{x}_{\mathrm{tm}}$] [\textit{tgt-lang}]: [$\mathbf{y}_{\mathrm{tm}}$] & \multirow{2}{*}{Yes} & [English]: [I have an apple.] [German]: [Ich habe einen Apfel.] & \multirow{2}{*}{65.54}\\ 
& [\textit{src-lang}]: [x] [\textit{tgt-lang}]: &  & [English]: [I have an orange.] [German]:&\\

\midrule

\multirow{1}{*}{15} & [\textit{src-lang}]: [x] [\textit{tgt-lang}]: & \multirow{1}{*}{No} & [English]: [I have an orange.] [German]: & \multirow{1}{*}{39.83}\\ 

\midrule

\multirow{2}{*}{16} & [\textit{src-lang}] = [$\mathbf{x}_{\mathrm{tm}}$] [\textit{tgt-lang}] = [$\mathbf{y}_{\mathrm{tm}}$] & \multirow{2}{*}{Yes} & [English] = [I have an apple.] [German] = [Ich habe einen Apfel.] & \multirow{2}{*}{66.45}\\ 
& [\textit{src-lang}] = [x] [\textit{tgt-lang}] = &  & [English] = [I have an orange.] [German] = &\\

\midrule

\multirow{2}{*}{17} & [\textit{src-lang}]=[$\mathbf{x}_{\mathrm{tm}}$] [\textit{tgt-lang}]=[$\mathbf{y}_{\mathrm{tm}}$] & \multirow{2}{*}{Yes} & [English]=[I have an apple.] [German]=[Ich habe einen Apfel.] & \multirow{2}{*}{\textbf{66.90}}\\ 
& [\textit{src-lang}]=[x] [\textit{tgt-lang}]= &  & [English]=[I have an orange.] [German]= &\\

\midrule

\multirow{1}{*}{18} & [\textit{src-lang}]=[x] [\textit{tgt-lang}]= & \multirow{1}{*}{No} & [English]=[I have an orange.] [German]= & \multirow{1}{*}{38.89}\\ 

\midrule

\multirow{2}{*}{19} & \{\textit{src-lang}\}=\{$\mathbf{x}_{\mathrm{tm}}$\} \{\textit{tgt-lang}\}=\{$\mathbf{y}_{\mathrm{tm}}$\} & \multirow{2}{*}{Yes} & \{English\}=\{I have an apple.\} \{German\}=\{Ich habe einen Apfel.\} & \multirow{2}{*}{65.48}\\ 
& \{\textit{src-lang}\}=\{x\} \{\textit{tgt-lang}\}= &  & \{English\}=\{I have an orange.\} \{German\}= &\\

\midrule

\multirow{2}{*}{20} & \{[\textit{src-lang}]=[$\mathbf{x}_{\mathrm{tm}}$]\} \{[\textit{tgt-lang}]=[$\mathbf{y}_{\mathrm{tm}}$]\} & \multirow{2}{*}{Yes} & \{[English]=[I have an apple.]\} \{[German]=[Ich habe einen Apfel.]\} & \multirow{2}{*}{63.32}\\ 
& \{[\textit{src-lang}]=[x]\} \{[\textit{tgt-lang}]=\} &  & \{[English]=[I have an orange.]\} \{[German]=\} &\\
\bottomrule
\end{tabular}
}
 \caption{Comparison of prompt templates in one-shot TM (i.e., $k=1$). Abbreviations are same as Figure \ref{fig:f1}.}
 \label{tab:app.diff_prompt}
\end{table*}

\section{More Experimental Results}
\label{app:Other experimental results}

\subsection{Evaluation by COMET-22}
\label{app:Evaluation by COMET-22}
\begin{table*}[t!]
\small
    \centering

 \resizebox{1.0\linewidth}{!}{
    \begin{tabular}{cc|cc|cc|ccc}
        \toprule
        \multicolumn{2}{c|}{\multirow{3}{*}[-0.1ex]{\normalsize{\textbf{Data}}}} & \multicolumn{2}{c|}{\normalsize{\textbf{WMT19 200M}}} & \multicolumn{2}{c|}{\normalsize{\textbf{WMT21 4B}}} & \multicolumn{3}{c}{\normalsize{\textbf{davinci-003 175B}}} \\
        & & \multicolumn{1}{c}{\multirow{2}{*}[-0.1ex]{\makecell{\textbf{NMT}}}} & \multicolumn{1}{c|}{\multirow{2}{*}[-0.1ex]{\makecell{\textbf{NMT+TM}}}} & \multicolumn{1}{c}{\multirow{2}{*}[-0.1ex]{\makecell{\textbf{NMT}}}} & \multicolumn{1}{c|}{\multirow{2}{*}[-0.1ex]{\makecell{\textbf{NMT+TM}}}} & \multicolumn{1}{c}{\multirow{2}{*}[0.4ex]{\textbf{LLM}}} &
        \multicolumn{1}{c}{\multirow{2}{*}[0.4ex]{\textbf{LLM+TM}}}  & \multicolumn{1}{c}{\multirow{2}{*}[0.4ex]{\textbf{LLM+TM}}}\\    
        & & & & & & \multicolumn{1}{c}{\multirow{1}{*}[0.1ex]{\footnotesize{(zero-shot)}}} &\multicolumn{1}{c}{\multirow{1}{*}[0.1ex]{\footnotesize{(one-shot)}}} & \multicolumn{1}{c}{\multirow{1}{*}[0.1ex]{\footnotesize{(few-shot)}}} \\
        \midrule
        \multirow{2}{*}{DGT-TM} & de $\rightarrow$ en & 85.99 & 87.28$_{(+1.29)}$ & 87.10 & 89.28$_{(+2.18)}$ & 83.86 & \multicolumn{1}{l}{88.74$_{(+4.88)}$} & \multicolumn{1}{l}{\textbf{89.47}$_{(+5.61)}$}\\   
        & en $\rightarrow$ de & 85.52 & 86.91$_{(+1.39)}$ & 86.89 & 89.01$_{(+2.12)}$ & 82.24 & \multicolumn{1}{l}{88.52$_{(+6.28)}$} & \multicolumn{1}{l}{\textbf{89.44}$_{(+7.20)}$} \\
        \midrule
        \multirow{2}{*}{JRC-A} & de $\rightarrow$ en & 85.85 & 85.80$_{(-0.05)}$ & 86.68 &  87.70$_{(+1.02)}$ & 84.15 & 87.79$_{(+3.64)}$ & \multicolumn{1}{l}{\textbf{88.46}$_{(+4.31)}$} \\
        & en $\rightarrow$ de & 86.53 & 86.25$_{(-0.28)}$ & 87.39 & \textbf{88.88}$_{(+1.49)}$ & 84.03 & \multicolumn{1}{l}{88.20$_{(+4.17)}$} & \multicolumn{1}{l}{88.84$_{(+4.81)}$} \\
        \bottomrule
    \end{tabular} 
    }

    \caption{COMET-22 scores of NMT models and LLMs on the DGT-TM and JRC-A dataset.}

    \label{tab:app:main_experiments_by_COMET-22}
    
\vspace{-1.1em}

\end{table*}
\begin{table*}[t!]
\small
    \centering

 \resizebox{1.0\linewidth}{!}{
    \begin{tabular}{cc|cc|cc|ccc}
        \toprule
        \multicolumn{2}{c|}{\multirow{3}{*}[-0.1ex]{\makecell{\textbf{Domain}}}} & \multicolumn{2}{c|}{\normalsize{\textbf{WMT19 200M}}} & \multicolumn{2}{c|}{\normalsize{\textbf{WMT21 4B}}} & \multicolumn{3}{c}{\normalsize{\textbf{davinci-003 175B}}} \\
        & & \multicolumn{1}{c}{\multirow{2}{*}[-0.1ex]{\makecell{\textbf{NMT}}}} & \multicolumn{1}{c|}{\multirow{2}{*}[-0.1ex]{\makecell{\textbf{NMT+TM}}}} & \multicolumn{1}{c}{\multirow{2}{*}[-0.1ex]{\makecell{\textbf{NMT
        }}}} & \multicolumn{1}{c|}{\multirow{2}{*}[-0.1ex]{\makecell{\textbf{NMT+TM}}}} & \multicolumn{1}{c}{\multirow{2}{*}[0.4ex]{\textbf{LLM}}} &
        \multicolumn{1}{c}{\multirow{2}{*}[0.4ex]{\textbf{LLM+TM}}}  & \multicolumn{1}{c}{\multirow{2}{*}[0.4ex]{\textbf{LLM+TM}}}\\
        & & & & & & \multicolumn{1}{c}{\multirow{1}{*}[0.1ex]{\footnotesize{(zero-shot)}}} &\multicolumn{1}{c}{\multirow{1}{*}[0.1ex]{\footnotesize{(one-shot)}}} & \multicolumn{1}{c}{\multirow{1}{*}[0.1ex]{\footnotesize{(few-shot)}}} \\
        \midrule
        \multicolumn{2}{c|}{\multirow{1}{*}{IT}}   &  83.04 & 83.87$_{(+0.83)}$ & 83.54 & \multicolumn{1}{l|}{85.09$_{(+1.55)}$}& 72.44 & 86.05$_{(+13.61)}$ & \textbf{87.27}$_{(+14.83)}$\\
        \multicolumn{2}{c|}{\multirow{1}{*}{Medical}}   &  83.30 & 83.61$_{(+0.31)}$ & 84.92 & 84.97$_{(+0.05)}$ & 80.76 & \multicolumn{1}{l}{84.97$_{(+4.21)}$} & \multicolumn{1}{l}{\textbf{86.63}$_{(+5.87)}$} \\
        \multicolumn{2}{c|}{\multirow{1}{*}{Koran}} &   72.42 & 72.00$_{(-0.42)}$ & \textbf{75.09} & 72.23$_{(-2.86)}$ & 73.35 & \multicolumn{1}{l}{73.65$_{(+0.30)}$} & \multicolumn{1}{l}{74.34$_{(+0.99)}$} \\
        \multicolumn{2}{c|}{\multirow{1}{*}{Law}} &  85.80 & 85.53$_{(-0.27)}$ & 86.75 & 87.23$_{(+0.48)}$ & 83.30 & \multicolumn{1}{l}{87.49$_{(+4.19)}$} & \multicolumn{1}{l}{\textbf{88.47}$_{(+5.17)}$} \\

        \bottomrule
    \end{tabular}
    }

    \caption{COMET-22 scores of NMT models and LLMs on the multi-domain dataset.}
    \label{tab:app:multi_domains_by_COMET-22}

\vspace{-0.9em}

\end{table*} 

Except for the BLEU scores, we also provide the COMET-22 scores as seen in Table \ref{tab:app:main_experiments_by_COMET-22} and Table \ref{tab:app:multi_domains_by_COMET-22}. We can see that despite LLM's poor performance on zero-shot, prompting LLM with a few TMs can achieve significant improvement. On the other hand, the few-shot learning+LLM system can still outperform the strong NMT+TM baseline in most cases.

\subsection{Performance of Different Prompt Templates}
\label{app:dif_prompt_performance}

In order to explore the effect of using different prompt templates on the performance of \texttt{davinci-003}, we use 20 prompt templates in the de $\rightarrow$ en direction of the DGT-TM dataset for experiments. Seen from table \ref{tab:app.diff_prompt}, the code-style template is better than the instruction-style template in most cases.

\begin{figure}
    \centering
    \begin{tikzpicture}[scale=0.9]
    \begin{axis}[
        legend entries={ en2de, de2en},
        legend columns=2,
        xlabel={$k$},
        ylabel={BLEU},
        ymajorgrids=true,
        xmajorgrids=true,
        grid style=dashed,
        legend style={draw=none,
        line width=1pt,
        at={(0.5,1.0)},
        anchor=south},
        xtick=data,
        yticklabel style={/pgf/number format/fixed,/pgf/number format/fixed zerofill,/pgf/number format/precision=2},
    ]
    \addplot+[color=kleinblue,smooth,mark size=1.5pt,mark=square*,mark options={line width=3.0pt}] 
    coordinates {(2,60.04)(3,61.29)(4,61.7)(5,62.02)(6,62.28)(7,62.25)(8,62.75)(9,62.7)};
    \addplot+[color=red,smooth,mark size=1.5pt,mark=triangle*,mark options={line width=3.0pt}]
    coordinates {(2,68.35)(3,69.1)(4,69.65)(5,69.9)(6,70.28)(7,70.64)(8,70.69)(9,70.63)};
    \end{axis}
\end{tikzpicture}
    \caption{BLEU scores of different $k$ on the DGT dataset}
    \label{fig:topk}
\end{figure}
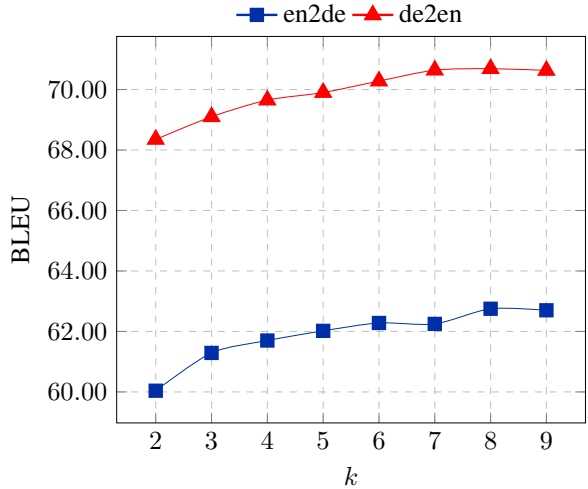

\begin{table*}[t!]
\centering
\begin{tabular}{c|cc|cc|cc|cc}
\toprule
\textbf{Lang} & \multicolumn{2}{c|}{\textbf{Cs-En}} & \multicolumn{2}{c|}{\textbf{Cs-It}} & \multicolumn{2}{c|}{\textbf{De-En}} & \multicolumn{2}{c}{\textbf{De-Fr}} \\ 
\textbf{Direction} & $\rightarrow$ & $\leftarrow$ & $\rightarrow$ & $\leftarrow$ & $\rightarrow$ & $\leftarrow$ & $\rightarrow$ & $\leftarrow$ \\
\midrule
w/o TM & 58.58 & 52.58 & 47.93 & 47.22 & 59.74 & 53.56 & 53.52 & 50.08  \\
w/ TM & 37.50 & 28.02 & 27.02 & 22.85 & 36.34 & 28.74 & 34.78 & 28.93  \\
$\Delta$ & +21.08 & +24.56 & +20.91 & +24.37 & +23.40 & +24.82 & +18.74 & +21.15  \\
\midrule
\textbf{Lang} & \multicolumn{2}{c|}{\textbf{Es-En}} & \multicolumn{2}{c|}{\textbf{Fr-En}} & \multicolumn{2}{c|}{\textbf{It-En}} & \multicolumn{2}{c}{\textbf{Ro-En}} \\
\textbf{Direction} & $\rightarrow$ & $\leftarrow$ & $\rightarrow$ & $\leftarrow$ & $\rightarrow$ & $\leftarrow$ & $\rightarrow$ & $\leftarrow$ \\
\midrule
w/o TM & 61.25 & 59.18 & 64.45 & 64.60 & 61.80 & 57.71 & 59.66 & 50.18  \\
w/ TM & 41.89 & 37.10 & 44.06 & 43.78 & 41.13 & 33.53 & 41.37 & 27.06  \\
$\Delta$ & +19.36 & +22.08 & +20.39 & +20.78 & +20.67 & +24.18 & +18.29 & +23.12  \\
\bottomrule
\end{tabular}
 \caption{Experiment results on 8 language-pairs from JRC-A.}
 \label{tab:app.JRC_score}
\end{table*}

\begin{table*}[t!]
\scalebox{0.80}{
\centering
\begin{tabular}{cc|c|ccccccccccc}
\toprule
\multicolumn{2}{c|}{\multirow{2}{*}{\textbf{Data}}} & \textbf{NMT} & \multicolumn{11}{c}{\textbf{FMS}} \\ 
& & \textbf{System} & 0 & 0.1 & 0.2 & 0.3 & 0.4 & 0.5 & 0.6 & 0.7 & 0.8 & 0.9 & 1.0 \\ 
\midrule
\multirow{3}{*}{DGT-TM} & en $\rightarrow$ de & \multirow{2}{*}{WMT19} & 57.39 & 57.53 & 57.70 & \textbf{58.02} & 57.71 & 57.08 & 56.07 & 54.73 & 53.05 & 48.67 & 37.06 \\
& de $\rightarrow$ en &  & \textbf{66.90} & \textbf{66.90} & 66.88 & 66.34 & 65.63 & 64.30 & 62.70 & 61.04 & 58.45 & 54.15 & 44.21 \\
\Xcline{2-14}{0.4pt}
& de $\rightarrow$ en & WMT21 & 66.90 & 66.90 & \textbf{67.06} & 67.02 & 66.74 & 65.84 & 64.58 & 63.08 & 60.59 & 57.33 & 49.21 \\
\midrule
\multirow{2}{*}{IT} & de $\rightarrow$ en & WMT19 & \textbf{47.46} & 44.48 & 43.85 & 43.30 & 40.78 & 36.92 & 35.78 & 32.55 & 29.27 & 28.17 & 26.65 \\
\Xcline{2-14}{0.4pt}
& de $\rightarrow$ en & WMT21 & \textbf{47.46} & 44.01 & 43.19 & 42.31 & 37.53 & 34.35 & 33.31 & 30.17 & 27.23 & 26.25 & 24.90 \\
\midrule
\multirow{2}{*}{Law} & de $\rightarrow$ en & WMT19 & \textbf{61.85} & 61.84 & 61.67 & 60.89 & 59.61 & 58.04 & 56.75 & 55.04 & 53.00 & 50.10 & 44.12 \\
\Xcline{2-14}{0.4pt}
& de $\rightarrow$ en & WMT21 & \textbf{61.85} & 61.83 & 62.00 & 61.99 & 61.39 & 60.34 & 59.32 & 57.98 & 56.31 & 53.93 & 47.56 \\
\midrule
\multirow{2}{*}{Medical} & de $\rightarrow$ en & WMT19 & \textbf{58.54} & 58.45 & 58.32 & 58.05 & 57.25 & 55.34 & 54.06 & 52.62 & 50.03 & 47.01 & 38.85 \\
\Xcline{2-14}{0.4pt}
& de $\rightarrow$ en & WMT21 & \textbf{58.54} & 58.45 & 58.32 & 58.11 & 57.30 & 55.44 & 54.14 & 52.74 & 20.15 & 47.14 & 38.97 \\
\bottomrule
\end{tabular}
}
 \caption{Performance of replacing the low-matching part of TMs at different thresholds of FMS with the translation results from NMT. For example, FMS 0.2 in first row means that TMs with FMS less than 0.2 are replaced by NMT translation results. }
 \label{tab:app.nmt-tm}
\end{table*}
\subsection{Experiments on More languages}
\label{app:Multi_languages_Complete}
We perform multi-lingual experiments on the JRC-A dataset, and in these experiments, we use the instruction-style template as shown in Figure \ref{fig:f1}. Table \ref{tab:app.JRC_score} shows the complete experiment results for the multi-language experiment. Great BLEU improvements are obtained on these datasets.

\subsection{Impact of $k$}
\label{app:few_shot_experimental}
To explore the effect of $k$ on the performance of \texttt{davinci-003} in the few-shot experiments, we conduct experiments with $k$ from 1 to 9 in both directions of the DGT-TM dataset. Figure \ref{fig:topk} shows a long-tail performance gain as $k$ increases.

\subsection{Impact of Orders of TM results}
To observe the effect of constructing the prompt template with different TMs similarity orders on the performance in the few-shot experiments, we constructed two types of prompt templates in the DGT-TM dataset with a few-shot sample size of 5. One is arranged in descending order of TMs similarity, and the TM adjacent to the sentence to be translated is the lowest similarity. The other one is arranged in ascending order of TMs similarity, and the TM adjacent to the sentence to be translated is the highest similarity. The results are shown in Table \ref{tab:app.performance_order_top5}.

\begin{table}[t!]
\centering
\begin{tabular}{c|cc}
\toprule
\textbf{Lang Direction} & \textbf{Descending} & \textbf{Ascending} \\ 
\midrule
de $\rightarrow$ en & 69.99 & 70.01 \\
en $\rightarrow$ de & 62.02 & 62.30 \\
\bottomrule
\end{tabular}
 \caption{The performance comparison of different templates which is constructed based on the similarity of TM when the number of few-shot samples is 5.}
 \label{tab:app.performance_order_top5}
\end{table}

\subsection{Performance on the WMT Datasets}
We conduct experiments on WMT14 en $\rightarrow$ de and WMT19 de $\rightarrow$ en directions. We use the same method as that used on the multi-domain dataset to process these two benchmarks. It is worth noting that the data obtained on these two benchmarks have a low similarity of TMs, as shown in Table \ref{tab:app.performance_wmt}. Table \ref{tab:app.performance_wmt} shows the performance of the LLM and baseline models on the WMT14 en $\rightarrow$ de and WMT19 de $\rightarrow$ en datasets.

\begin{table}[t!]
\small
\centering
\begin{tabular}{cc|cc}
\toprule
\multirow{2}{*}{\textbf{Model}} & \multirow{2}{*}{\textbf{TM}} & \textbf{WMT14} & \textbf{WMT19} \\ 
& & \textbf{En2De} & \textbf{De2En} \\ 
\midrule
\multirow{2}{*}{Transformer-base} & w/o TM & 27.59 & 39.67 \\
& w/ TM & 21.86 & 40.22 \\
\midrule
\multirow{2}{*}{text-davinci-003} & w/o TM & 29.58 & 40.85 \\
& w/ TM & 28.11 & 36.63 \\
\bottomrule
\end{tabular}
 \caption{Comparison of performance on WMT dataset.}
 \label{tab:app.performance_wmt}
\end{table}

\subsection{Performance of Different Sized Models}
\label{app:five_gpt_model}
Moreover, we conduct experiments using ``small'' models such as \texttt{text-curie-001} and \texttt{text-babbage-001}. But their performance is far away behind \texttt{davinci-003} whose outputs contain null in lines sometimes. We attribute this to the lack of emergent abilities of big models \cite{wei2022emergent}. The results are shown in Table \ref{tab:app.five_gpt_model}.

\begin{table}[t!]
\centering
\begin{tabular}{r|c}
\toprule
\multicolumn{1}{c|}{\textbf{Model}} & \textbf{BLEU} \\ 
\midrule
text-davinci-003 & 66.90 \\
davinci(GPT3) & 65.48 \\
text-curie-001 & 42.30 \\
text-babbage-001 & 37.72 \\
text-ada-001 & 14.65 \\
\bottomrule
\end{tabular}
 \caption{Comparison of performance with different size models on DGT-TM de $\rightarrow$ en.}
 \label{tab:app.five_gpt_model}
\end{table}

\end{document}